\documentclass[10pt,twocolumn,letterpaper]{article}

\usepackage{wacv}
\usepackage{times}
\usepackage{epsfig}
\usepackage{graphicx}
\usepackage{amsmath}
\usepackage{amssymb}

\usepackage{multirow}
\usepackage{fontawesome}
\usepackage{enumitem} 

\usepackage[pagebackref=true,breaklinks=true,letterpaper=true,colorlinks,bookmarks=false]{hyperref}

\def\wacvPaperID{537} 

\wacvfinalcopy 

\ifwacvfinal
\def\assignedStartPage{1} 
\fi

\ifwacvfinal
\usepackage[breaklinks=true,bookmarks=false]{hyperref}
\else
\usepackage[pagebackref=true,breaklinks=true,colorlinks,bookmarks=false]{hyperref}
\fi

\ifwacvfinal
\else
\fi

\begin{document}

\title{Danish Fungi 2020 -- Not Just Another Image Recognition Dataset}

\author{
Luk\'{a}\v{s} Picek \\ University of West Bohemia \\ {\tt\small picekl@kky.zcu.cz} 
\and Milan \v{S}ulc, Ji\v{r}\'{i} Matas \\ CTU in Prague \\ {\tt\small {{sulcmila,matas}@fel.cvut.cz}}
\and Thomas S. Jeppesen \\ GBIF \\ {\tt\small tsjeppesen@gbif.org}
\and Jacob Heilmann-Clausen, Thomas Læssøe, Tobias Frøslev \\ University of Copenhagen \\ {\tt\small jheilmann-clausen@snm.ku.dk, thomasl@bio.ku.dk, tobiasgf@sund.ku.dk}
}

\maketitle

\begin{abstract}

We introduce a novel fine-grained dataset and benchmark, the Danish Fungi 2020 (DF20). The dataset, constructed from observations submitted to the Atlas of Danish Fungi, is unique in its taxonomy-accurate class labels, small number of errors, highly unbalanced long-tailed class distribution, rich observation metadata, and well-defined class hierarchy. DF20 has zero overlap with ImageNet, allowing unbiased comparison of models fine-tuned from publicly available ImageNet checkpoints. The proposed evaluation protocol enables testing the ability to improve classification using metadata -- e.g. precise geographic location, habitat, and substrate, facilitates classifier calibration testing, and finally allows to study the impact of the device settings on the classification performance. 
Experiments using Convolutional Neural Networks (CNN) and the recent Vision Transformers (ViT) show that DF20 presents a challenging task. Interestingly, ViT achieves results superior to CNN baselines with 80.45\% accuracy and 0.743 macro F1 score, reducing the CNN error by 9\% and 12\% respectively. A simple procedure for including metadata into the decision process improves the classification accuracy by more than 2.95 percentage points, reducing the error rate by 15\%. The source code for all methods and experi- ments is available at \url{https://sites.google.com/view/danish-fungi-dataset}.

\end{abstract}
\section{Introduction}
Publicly available datasets and benchmarks accelerate machine learning research and allow for quantitative comparison of novel methods. In the area of deep learning and computer vision, the rapid progress over the past decade was, to a great extent, facilitated by the publication of large-scale image datasets. In the case of image recognition, the formation of the ImageNet\,\cite{imagenet} database and its usage in the ILSVRC\footnote{\,The ImageNet Large Scale Visual Recognition Challenge.}\,challenge\,\cite{ILSVRC15}, together with PASCAL\,VOC\,\cite{voc} among others, helped start the CNN revolution. The same holds for the problem of fine-grained visual categorization\,(FGVC), where datasets and challenges like PlantCLEF\,\cite{plantclef2016, plantclef2017, plantclef2015}, iNaturalist\,\cite{inaturalist2017}, CUB\,\cite{dataset-CUBS}, and Oxford Flowers \,\cite{dataset-flower}, have helped to develop and evaluate novel approaches to fine-grained domain adaptation\,\cite{domain_adap}, domain specific transfer learning\,\cite{transfer_learning}, image retrieval\,\cite{ft_imagenet_3, sohn2016improved, zhai2019classification}, unsupervised visual representation, few-shot learning \,\cite{wertheimer2019few}, transfer learning\,\cite{transfer_learning}, prior-shift\,\cite{sulc2018improving} and many others.

\begin{figure}[t!]
\begin{center}
\vspace{0.4cm}
\includegraphics[width=1.0\linewidth]{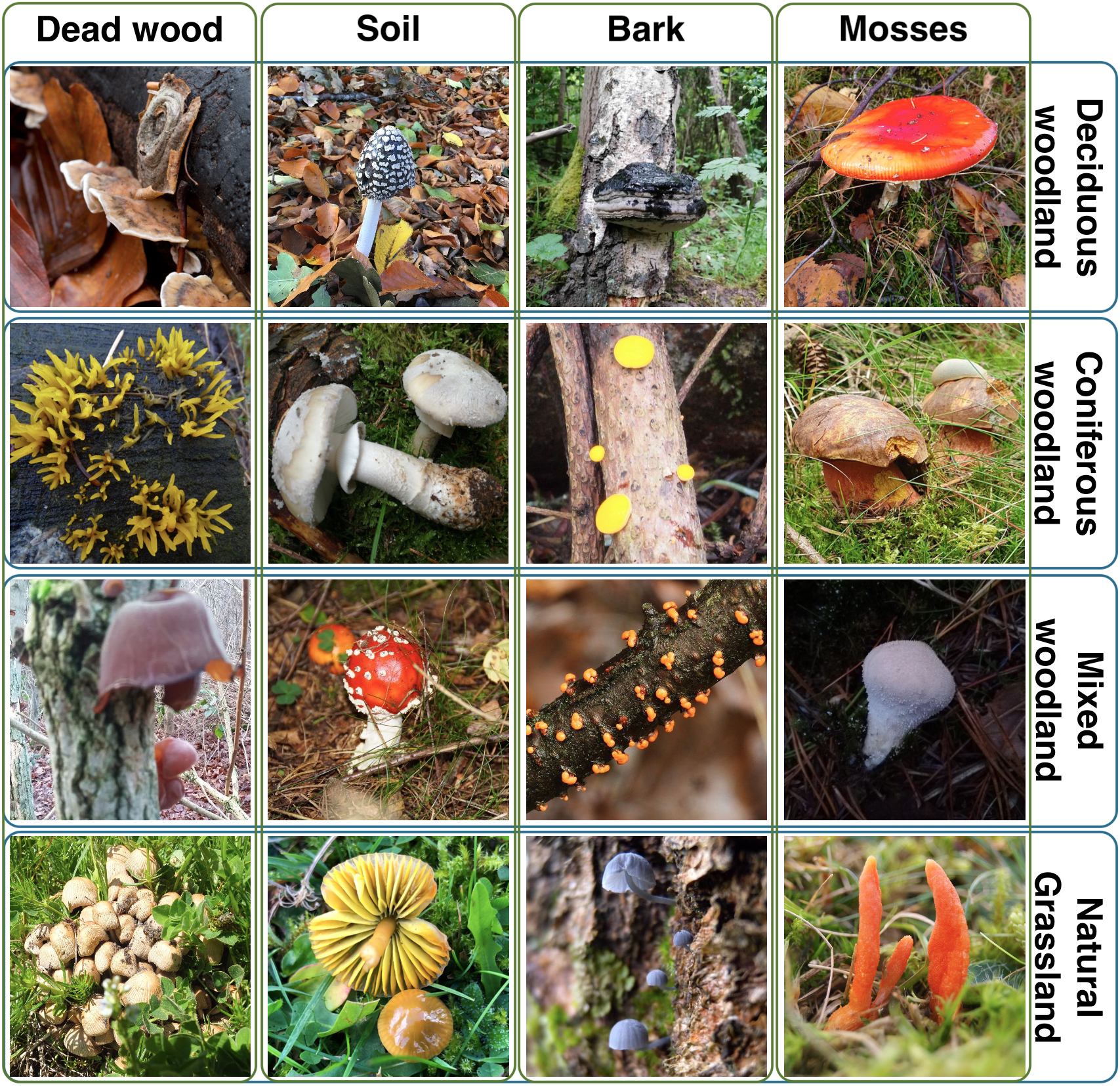}
\caption{Selected images from the \textbf{DF20} dataset from different Habitats (Rows) that grow on a variety of Substrates (Columns).}
\label{fig:genus_metadata}
\vspace{-0.5cm}
\end{center}
\end{figure}

While the datasets have been extremely useful for the image recognition community, there are issues that limit their relevance to real-world applications. We mention several such problems. Uniform class distribution, common in research datasets, are rare in practice. Often, class prior distributions are the same in the training and test splits. This is a standard machine learning assumption that, nevertheless, is not valid if the collection of training data differs from the deployment of the trained system, which is not rare. A non-negligible percentage of noisy-labels restricts quality assessment\,\cite{done_with_imagenet}, and, despite CNN's surprising robustness to label noise\,\cite{unreas_effec}, may influence the perceived relative merit of learning algorithms. Some commonly used datasets\,\cite{imagenet, dataset-Cars, dataset-flower} are saturated in accuracy or close to the point, leaving limited space for improvement in future research\,\cite{done_with_imagenet}. 
Extremely large dataset sizes might discourage researchers that do not have access to massive computational resources as experiments have become time and hardware demanding. \\
\begin{table}[b]
\small
\vspace{-0.2cm}
\begin{center}
\renewcommand{\arraystretch}{1.0}
\setlength{\tabcolsep}{0.4em}
\begin{tabular}{|l|r|r|r|}
\cline{2-4}
\multicolumn{1}{l|}{ } & ~~\#\,Classes & ~~\#\,Training & ~~\#\,Testing \\
\hline
FGVC-Aircafts\,\cite{aircafts}              &  102   &  6,732  &  3,468   \\
Standford Cars\,\cite{dataset-Cars}          &  196   &  8,144  &  8,041   \\
VMMRdb\,\cite{vmmrdb}                        &  9,170    &  291,752  &  $\times$   \\
\hline
Oxford Flowers\,\cite{dataset-flower}                 &    102 &   1,020 &   7,169  \\
Stanford Dogs\,\cite{dogsnap_dataset}                 &    120 &  12,000 &   8,580  \\
DogSnap\,\cite{dogsnap_dataset}                       &    133 &   4,776 &   3,575  \\
LeafSnap\,\cite{leafsnap}                             &    185 &  30,866 & $\times$ \\
CUB 200-2011\,\cite{dataset-CUBS}                     &    200 &   5,994 &   5,784  \\
VegFru\,\cite{vegfru_dataset}                         &    292 &  29,200 & 116,931  \\
Birdsnap\,\cite{dataset-birdsnap}                     &    500 &  47,386 &   2,443  \\
NABirds\,\cite{nabirds_dataset}                       &    555 &  48,562 & $\times$ \\
PlantCLEF 2015\,\cite{plantclef2015}                  &  1,000 &  91,758 &  21,446  \\
iNaturalist 2017\,\cite{inaturalist2017}              &  5,089 & 579,184 &  95,986  \\
PlantCLEF 2017{$^{\dagger}$}\,\cite{plantclef2017}~ & 10,000 & 230,658 &  25,629  \\
\hline \hline
\textbf{DF20 - Mini}        &    182 &  32,753 &   3,640  \\
\textbf{DF20}               &  1,604 & 266,344 &  29,594  \\
\hline
\end{tabular}
\end{center}
\caption{Overview of publicly available FGVC datasets, nature-related (middle section) and other (top), and the number of images and categories. {$^{\dagger}$}\,Images with ``clean'' (accurate) labels only.}
\label{table:fgvc_datasets}
\end{table}
With these observations in mind, we introduce {\bf the DF20 dataset} with a number of unique characteristics. Its class labels are exceptionally accurate, annotated by domain experts -- Mycologists with specialization on specific Families/Genera. Moreover, we are preparing private test with labels acquired by DNA sequencing. The minimal error levels allow highly accurate performance evaluation. 
With its zero overlap with ImageNet, it allows an unbiased comparison of models fine-tuned from publicly available ImageNet checkpoints.

The class frequencies in DF20 follow the natural species distribution, which is long-tailed. The frequencies change significantly within the calendar year, making the data suitable for testing the response of the classifier to differing long tailed distributions and changing class priors. 
The continuous data flow of collection over a long period provides a ground for modelling and exploiting the temporal phenomena on different scales, e.g., month, season, year.

The visual data is accompanied with metadata for more than 99\% of the image observations. The rich metadata includes information related to the environment, place, time and full taxonomy labels and enables testing the ability to improve classification accuracy using different metadata types -- time, precise location, habitat, and substrate type, to perform hierarchical classification, evaluate fine-grained classification on different levels of granularity (taxonomic ranks), to test classifier calibration, and to model intra-metadata and metadata-visual appearance relationships.
Moreover, EXIF metadata is available for many observations, which is useful, e.g., for studying the impact of the device settings on classification performance.

\textbf{The DF20 Benchmark.}
To allow evaluation at any time, we have prepared a web-based public automatic benchmark\,\footnote{\url{www.aicrowd.com/challenges/danish-fungi-2020}} for different scenarios, including visual-based, metadata focused or classifier-calibration related research.
Besides the full benchmark, we introduce DF20\,-\,mini, a small subset with roughly 1/10 of the data and species, for fast, low-energy friendly prototyping. DF20\,-\,mini includes six well-known genera of fungi forming fruit-bodies of the toadstool type, and offers, surprisingly, an even more challenging problem then the full benchmark, while having a compact size. 

We prepared a baseline performance evaluation, including the quantitative and qualitative analysis of the results for a number of well-known CNN and recent ViT architectures\,\cite{vit}. The recent ViT achieves excellent results in fine-grained classification outperforming the state-of-the-art CNN classifiers. We show that ViT performs way better on the FGVC domain, where attention to detail is needed, than in a common object recognition. 
We show that both the DF20 and DF20\,-\,Mini benchmarks are far from saturated as the best performing model - ViT-Large/16-384 - achieved 80.45\% and 75.85\% accuracy on DF20 and DF20\,-\,Mini, respectively.  
We propose a simple method for processing the habitat, substrate and time (month) metadata, showing that -- even with the simple approach -- utilizing the metadata increases the classification performance significantly. To support and accelerate future research on the DF20 we open source the code through the public GitHub.

\begin{figure*}[!t]
  \begin{center}
  \footnotesize
  \renewcommand{\arraystretch}{1.0}
  \setlength{\tabcolsep}{1.5pt}
  \begin{tabular}{|ccc|ccc|ccc|}
  \hline
    \textit{Russula} & \textit{Russula} & \textit{Russula} & \textit{Hortiboletus} & \textit{Suillellus} & \textit{Neoboletus} & \textit{Amanita} & \textit{Amanita} & \textit{Amanita} \\
    \noalign{\vskip -1mm}
    \textit{emetica} & \textit{paludosa} & \textit{rosea} & \textit{rubellus} & \textit{queletii} & \textit{luridiformis} & \textit{muscaria} & \textit{rubescens} & \textit{pantherina} \\
    \includegraphics[width=0.105\linewidth, height=0.105\linewidth]{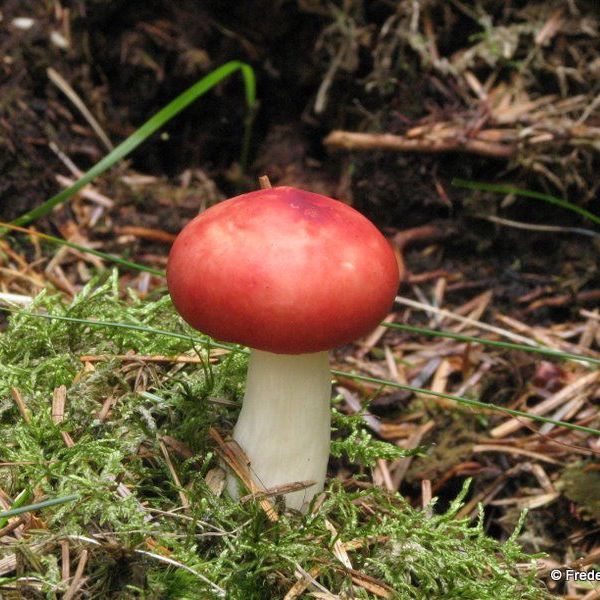} &
    \includegraphics[width=0.105\linewidth, height=0.105\linewidth]{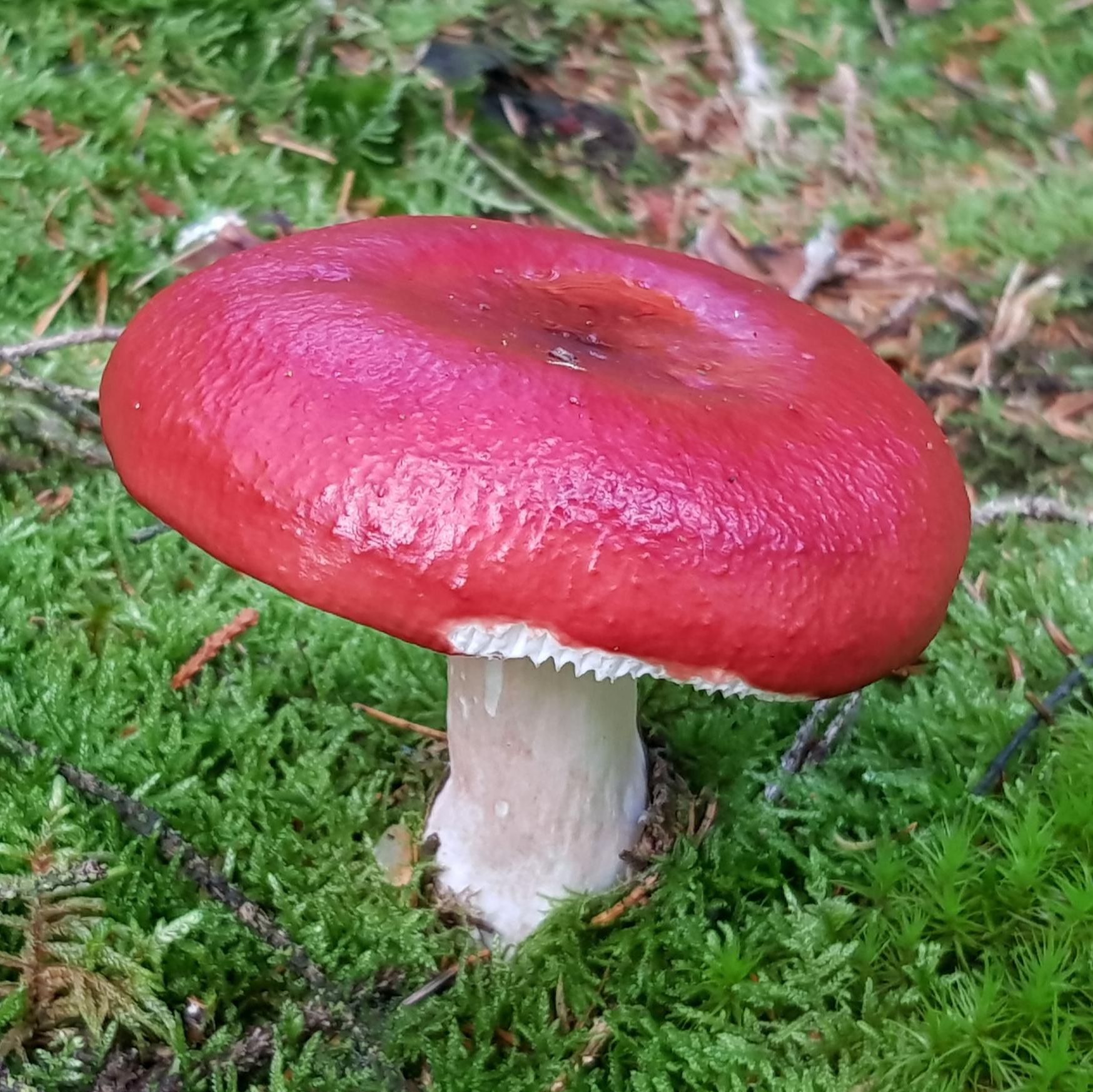} &
    \includegraphics[width=0.105\linewidth, height=0.105\linewidth]{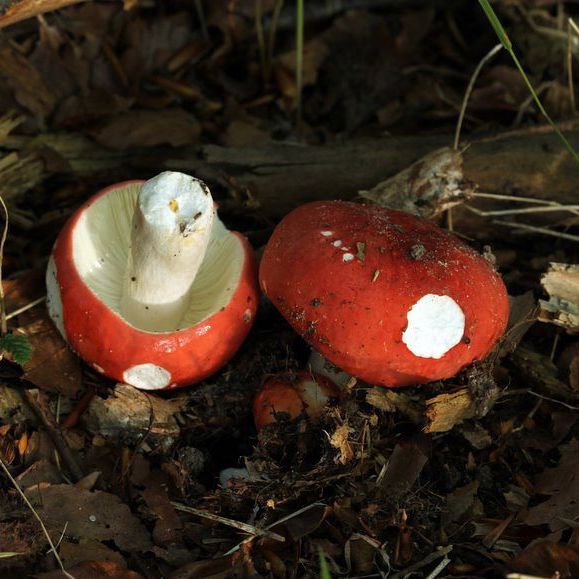} &
    \includegraphics[width=0.105\linewidth, height=0.105\linewidth]{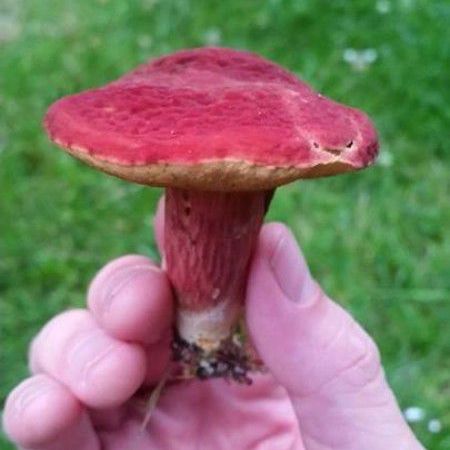} &
    \includegraphics[width=0.105\linewidth, height=0.105\linewidth]{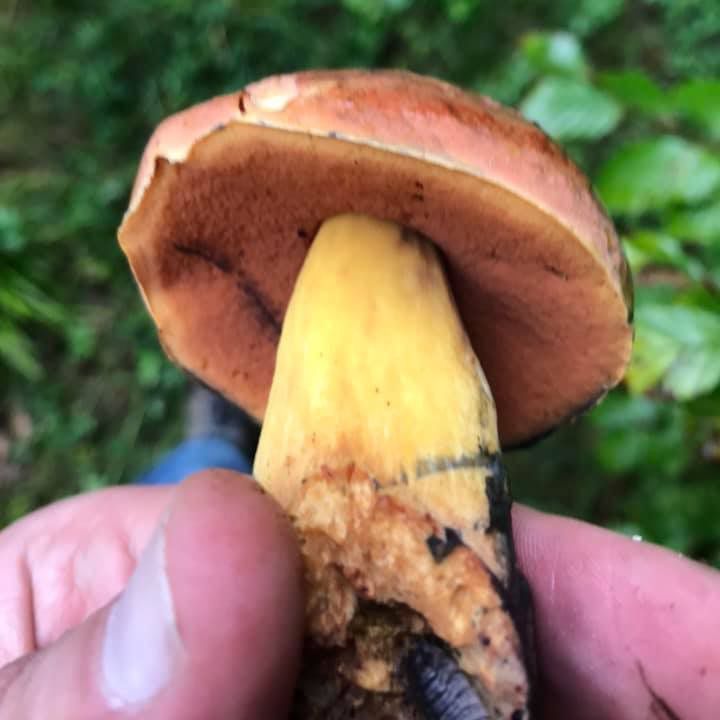} &
    \includegraphics[width=0.105\linewidth, height=0.105\linewidth]{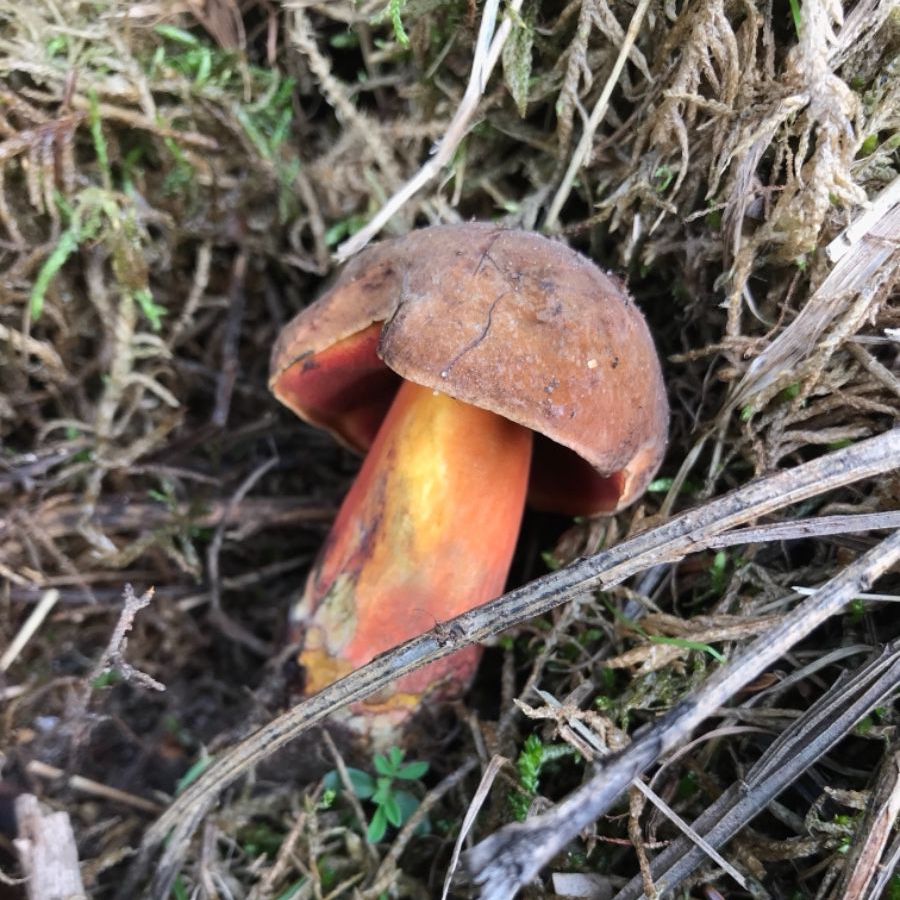} &
    \includegraphics[width=0.105\linewidth, height=0.105\linewidth]{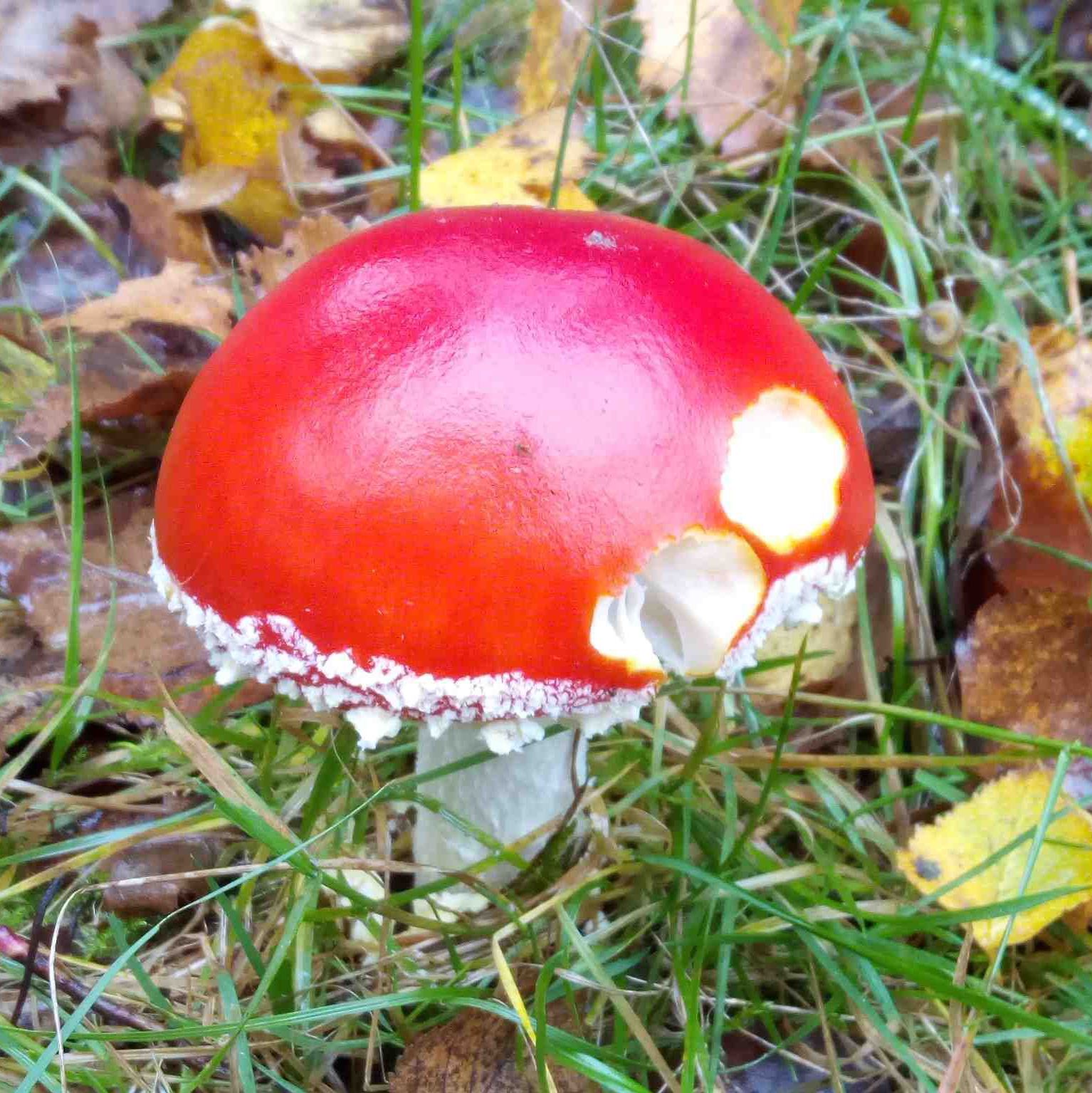} &
    \includegraphics[width=0.105\linewidth, height=0.105\linewidth]{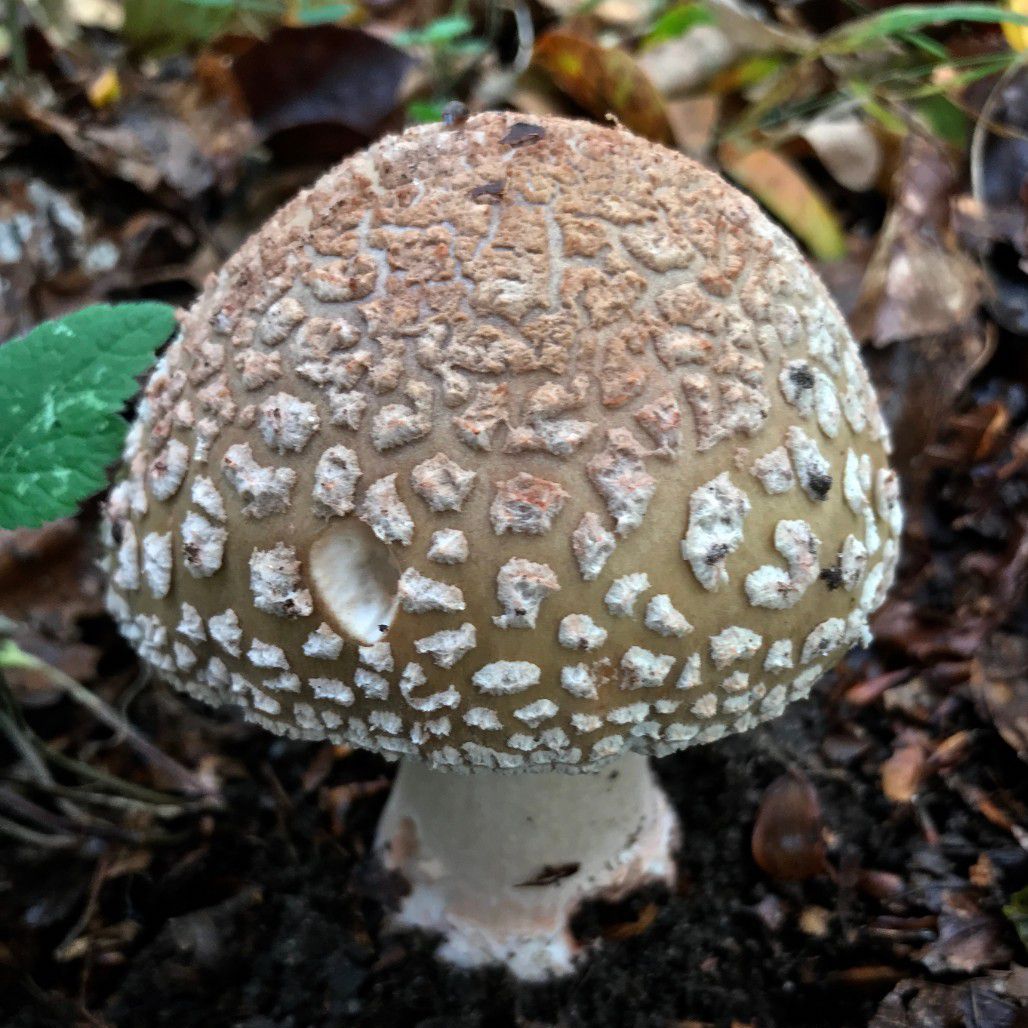} &
    \includegraphics[width=0.105\linewidth, height=0.105\linewidth]{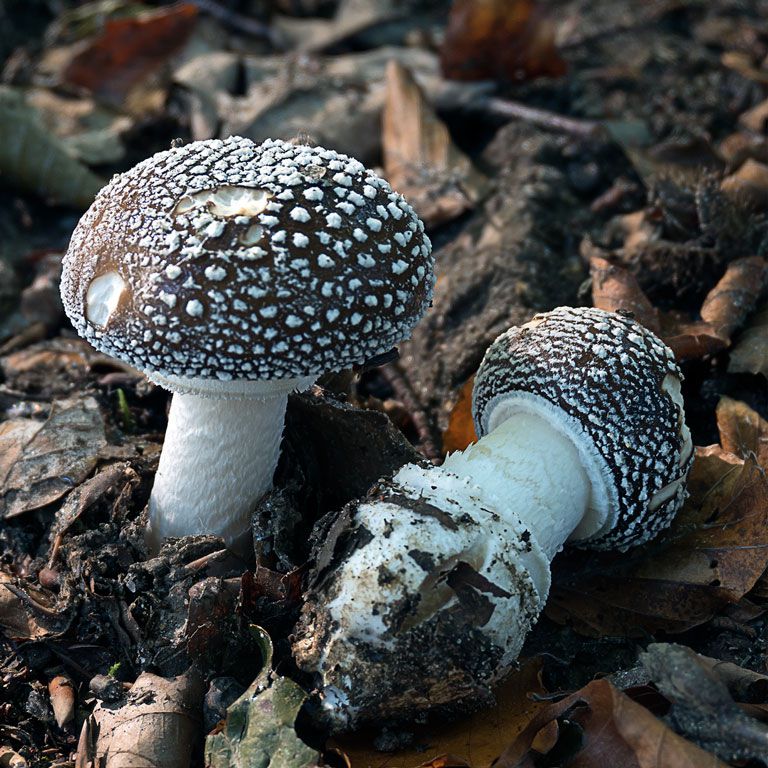} \\
    \includegraphics[width=0.105\linewidth, height=0.105\linewidth]{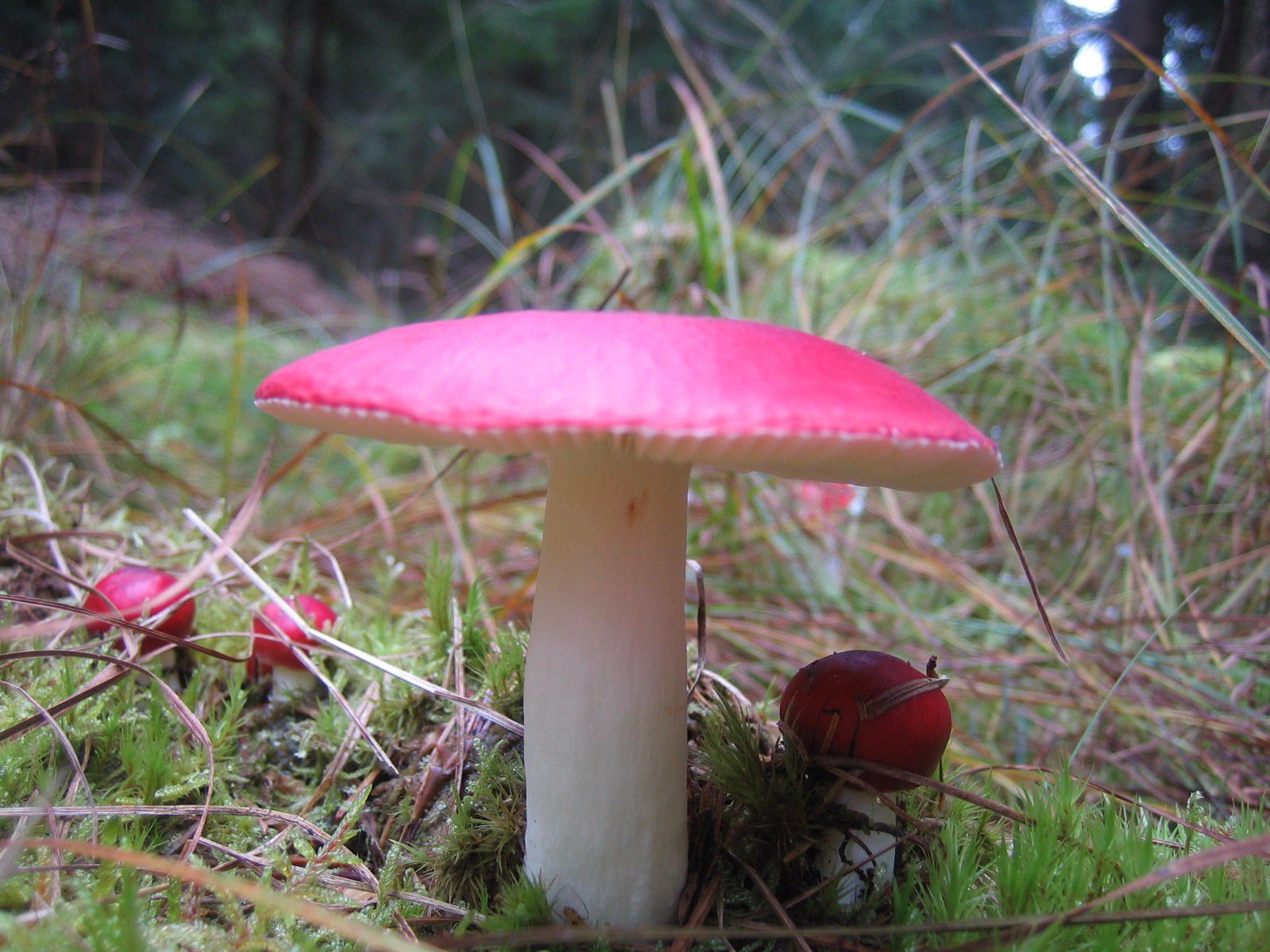} &
    \includegraphics[width=0.105\linewidth, height=0.105\linewidth]{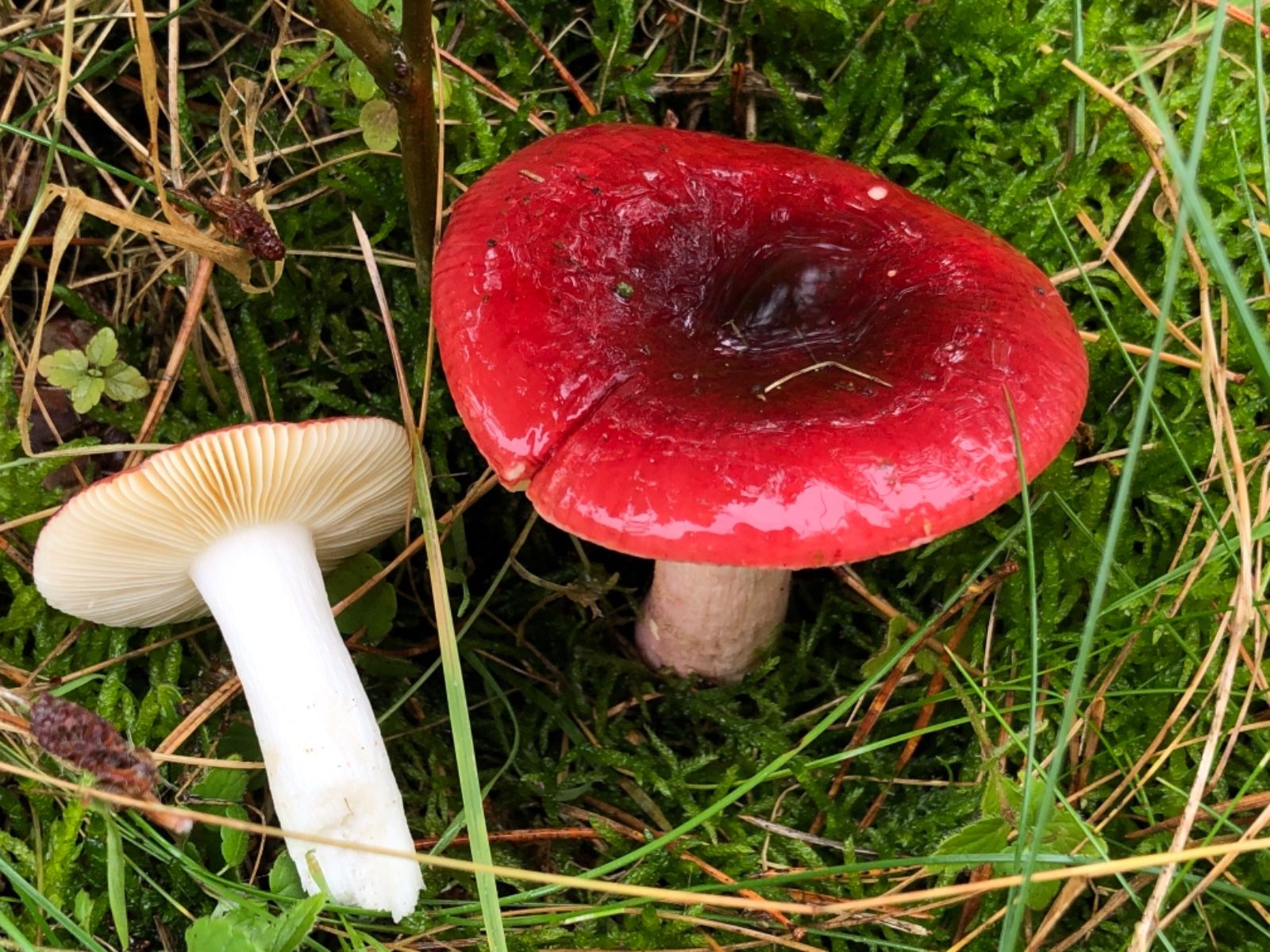} &
    \includegraphics[width=0.105\linewidth, height=0.105\linewidth]{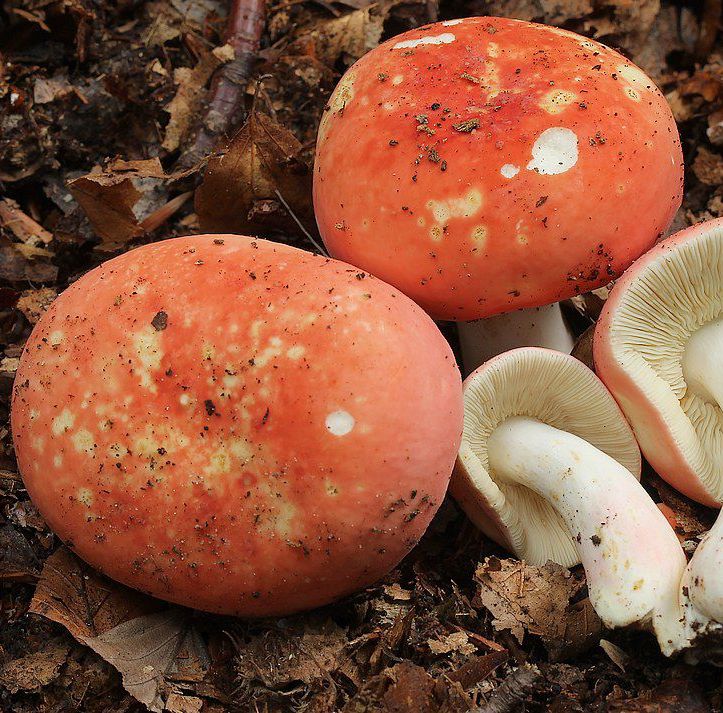} &
    \includegraphics[width=0.105\linewidth, height=0.105\linewidth]{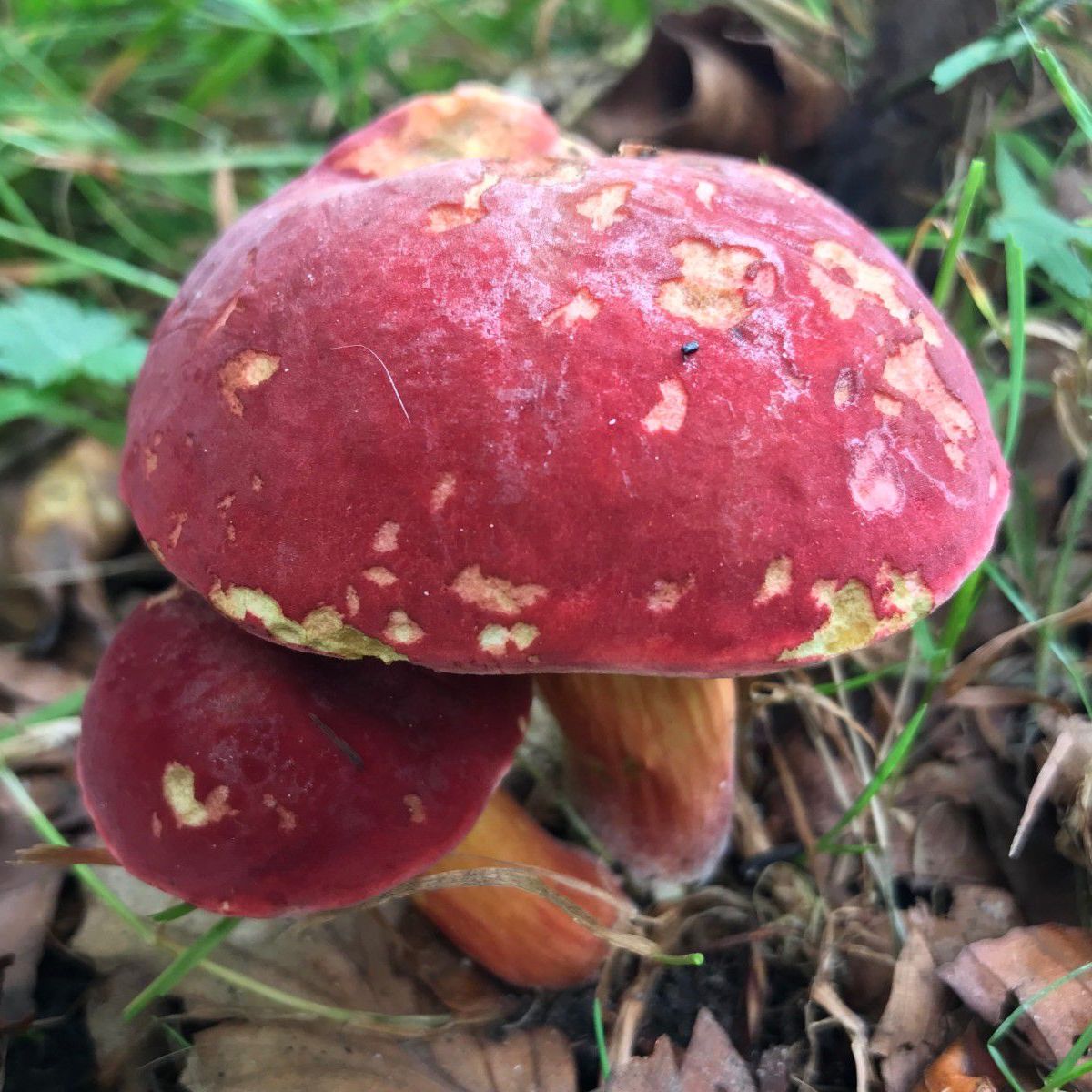} &
    \includegraphics[width=0.105\linewidth, height=0.105\linewidth]{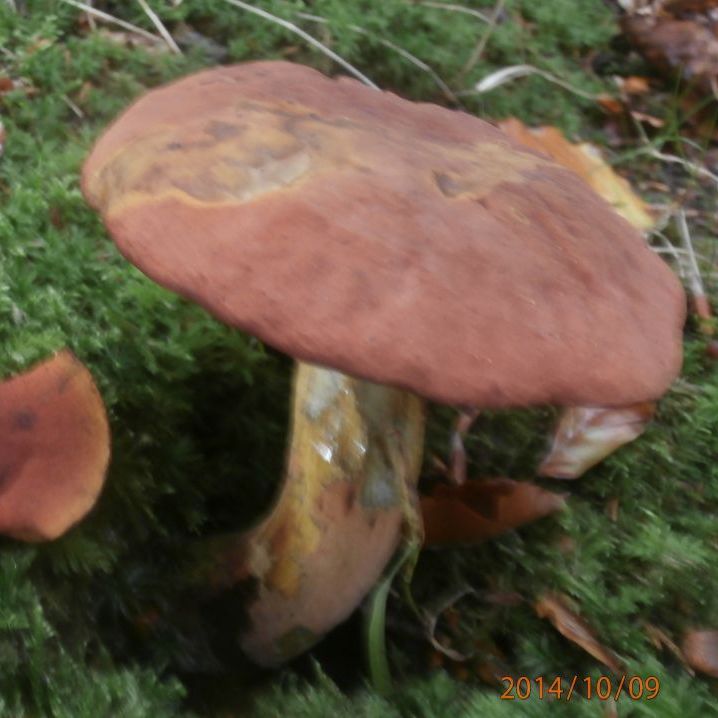} &
    \includegraphics[width=0.105\linewidth, height=0.105\linewidth]{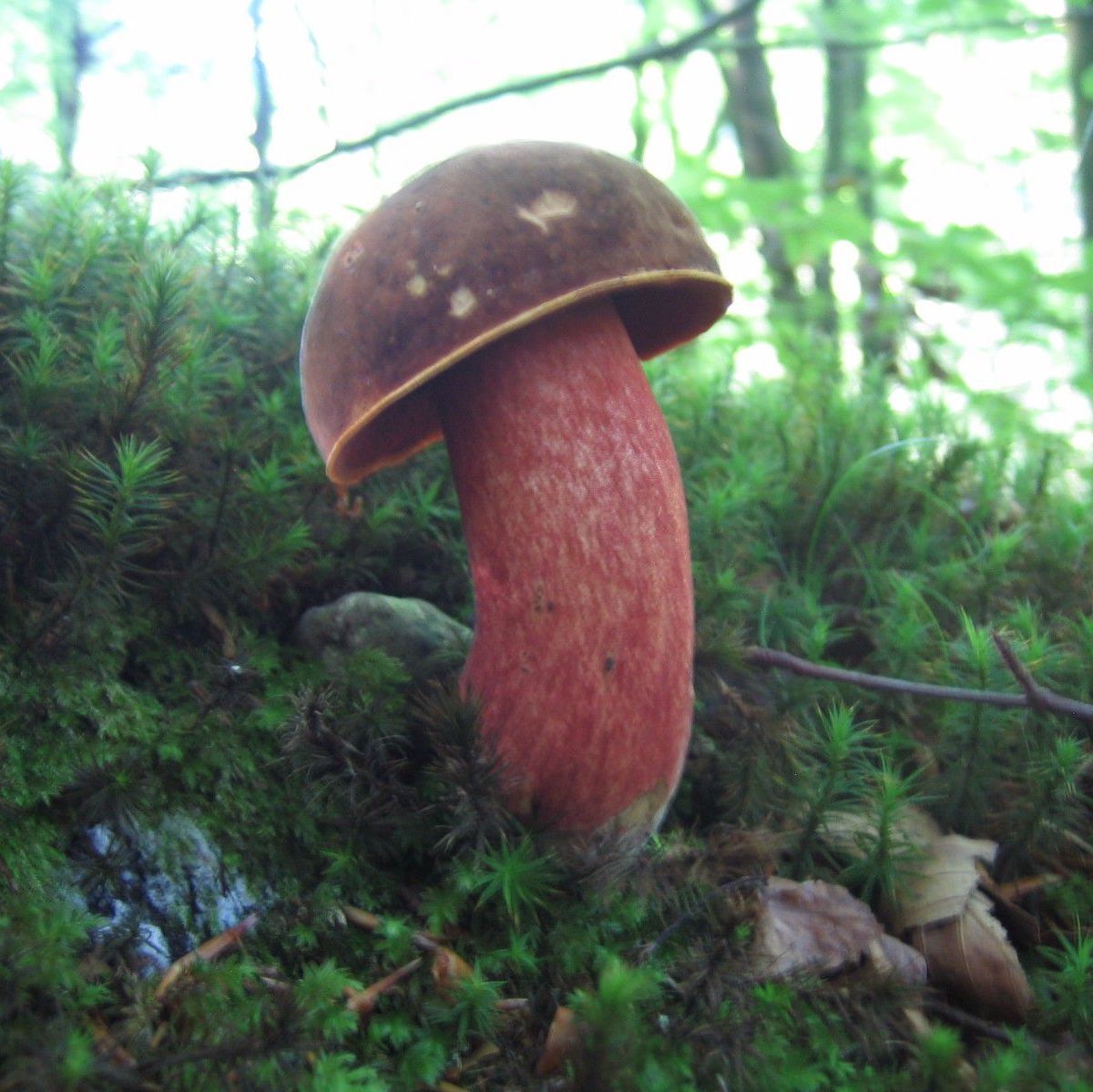} &
    \includegraphics[width=0.105\linewidth, height=0.105\linewidth]{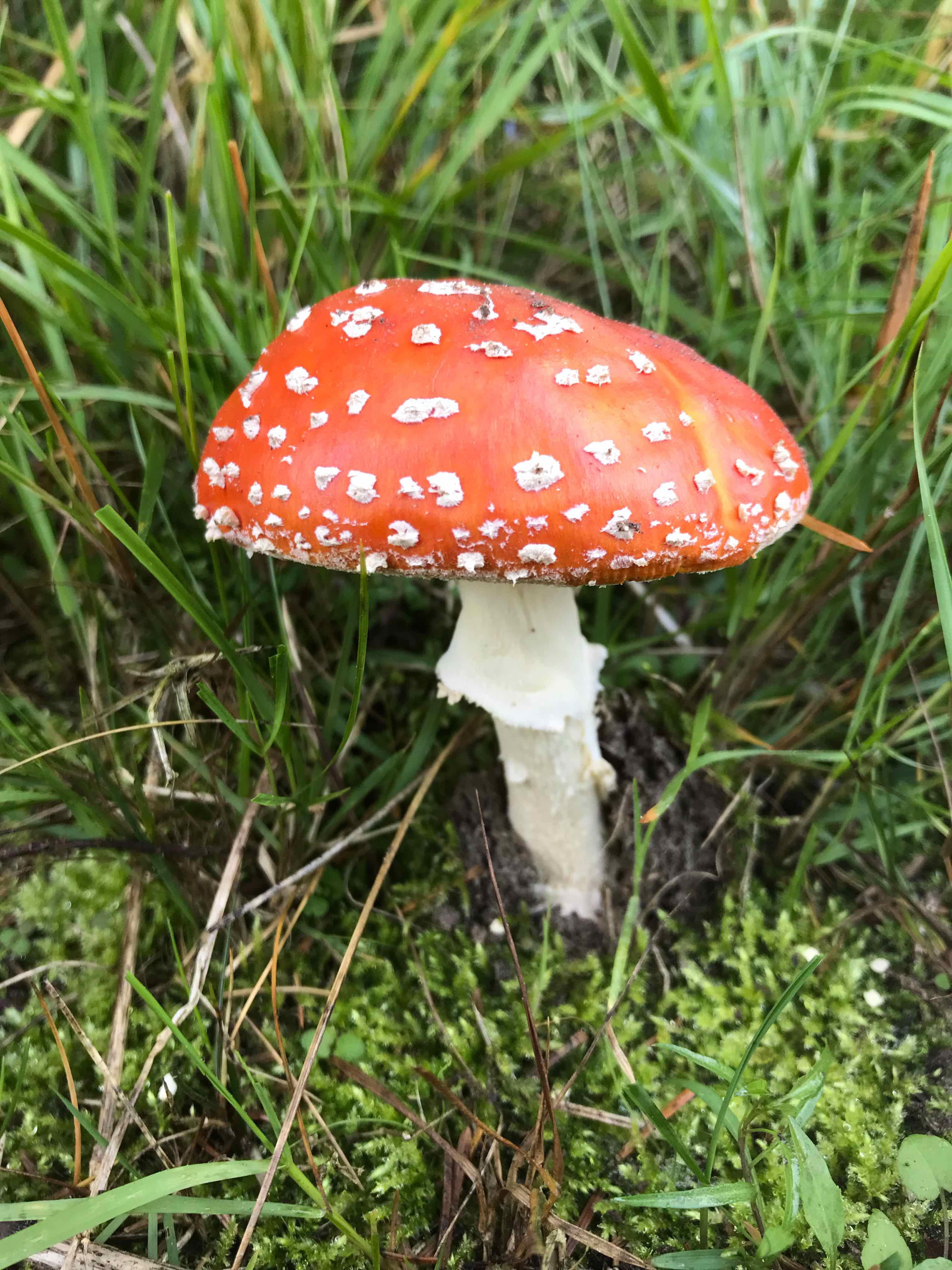} &
    \includegraphics[width=0.105\linewidth, height=0.105\linewidth]{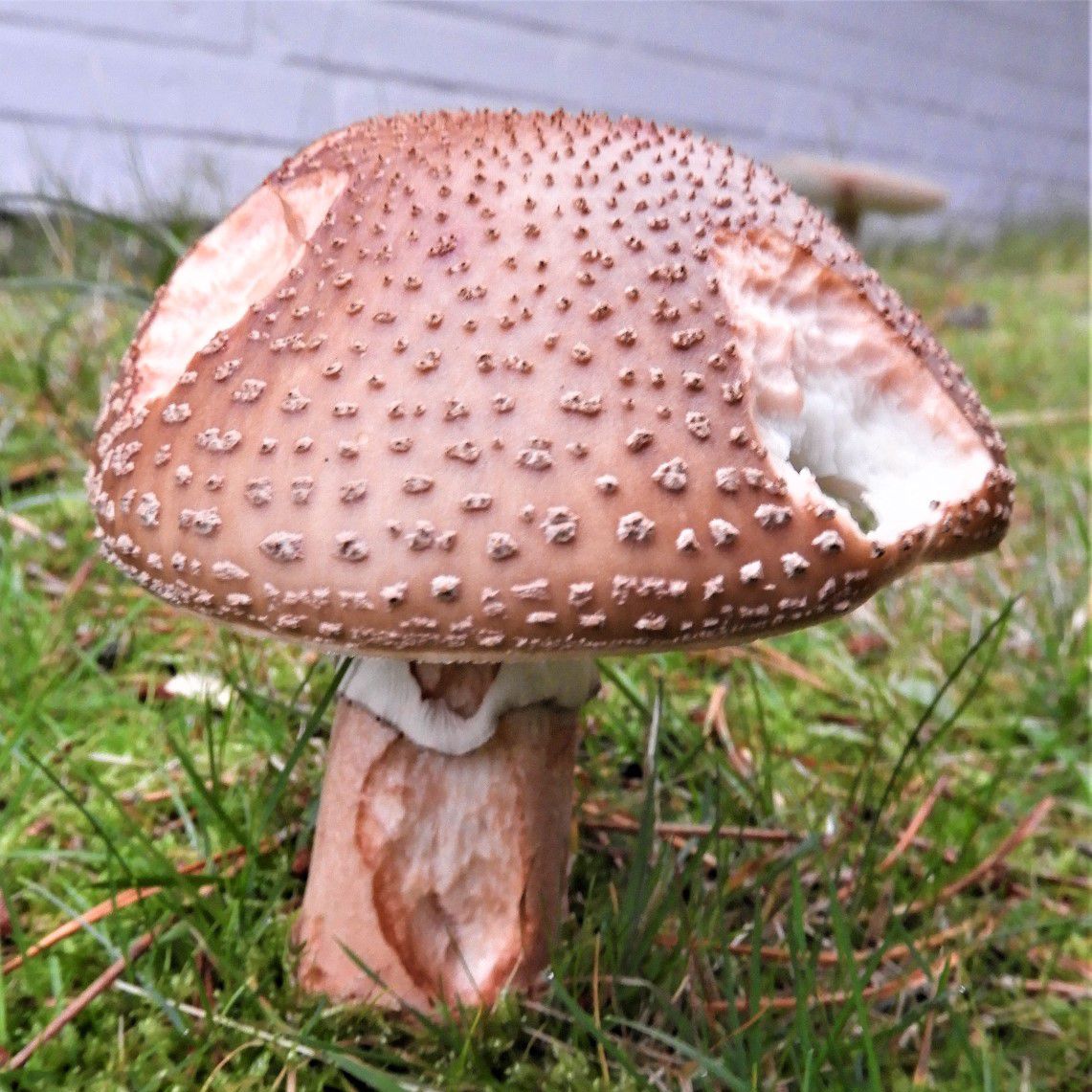} &
    \includegraphics[width=0.105\linewidth, height=0.105\linewidth]{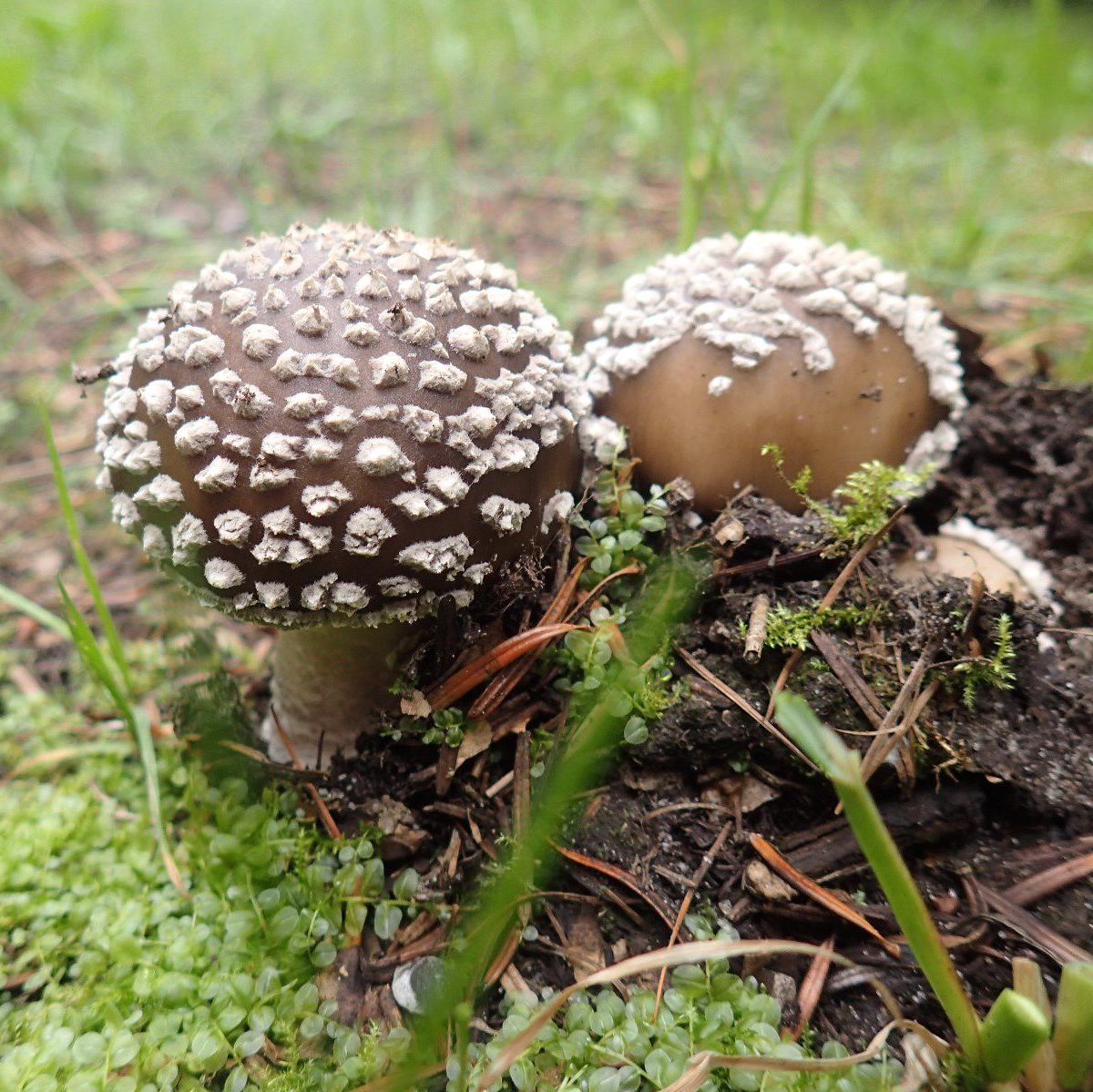} \\
    \includegraphics[width=0.105\linewidth, height=0.105\linewidth]{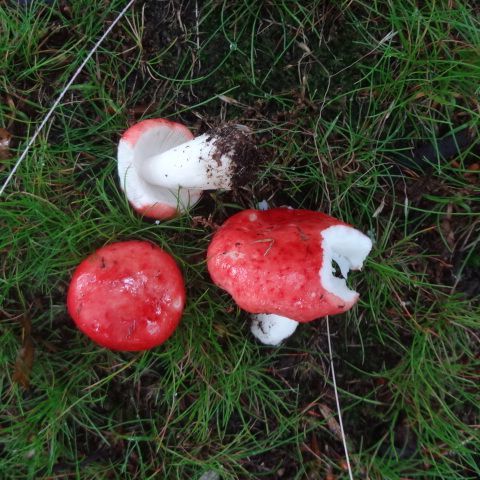} &
    \includegraphics[width=0.105\linewidth, height=0.105\linewidth]{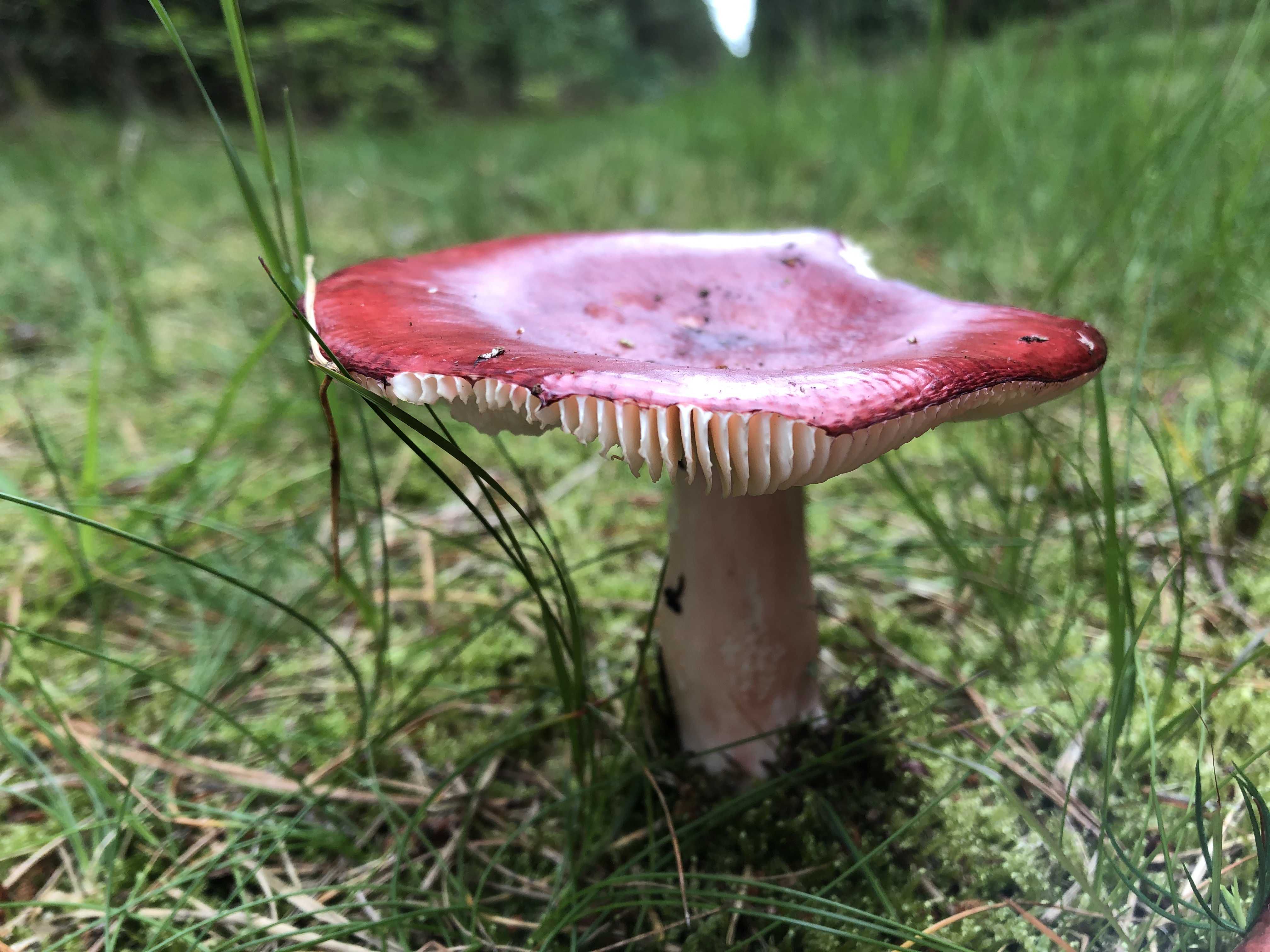} &
    \includegraphics[width=0.105\linewidth, height=0.105\linewidth]{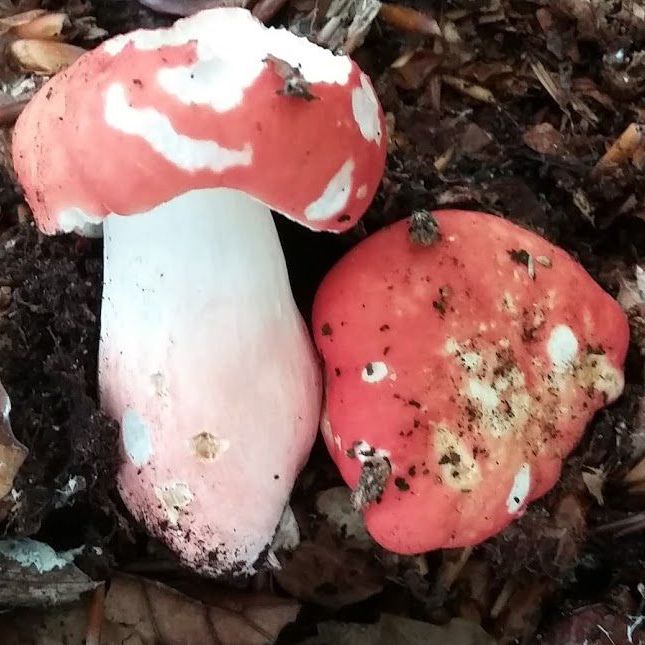} &
    \includegraphics[width=0.105\linewidth, height=0.105\linewidth]{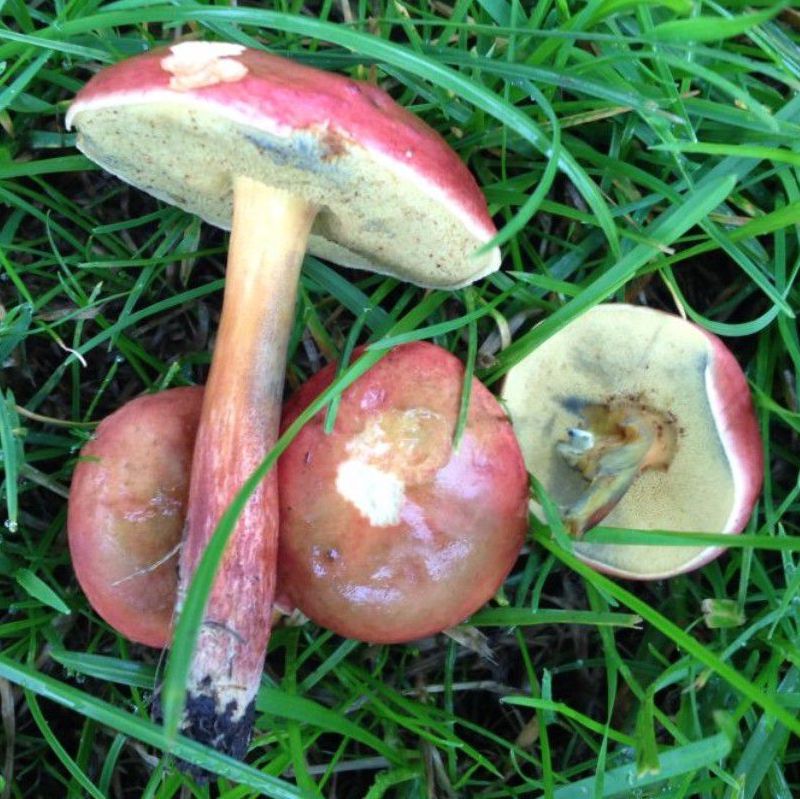} &
    \includegraphics[width=0.105\linewidth, height=0.105\linewidth]{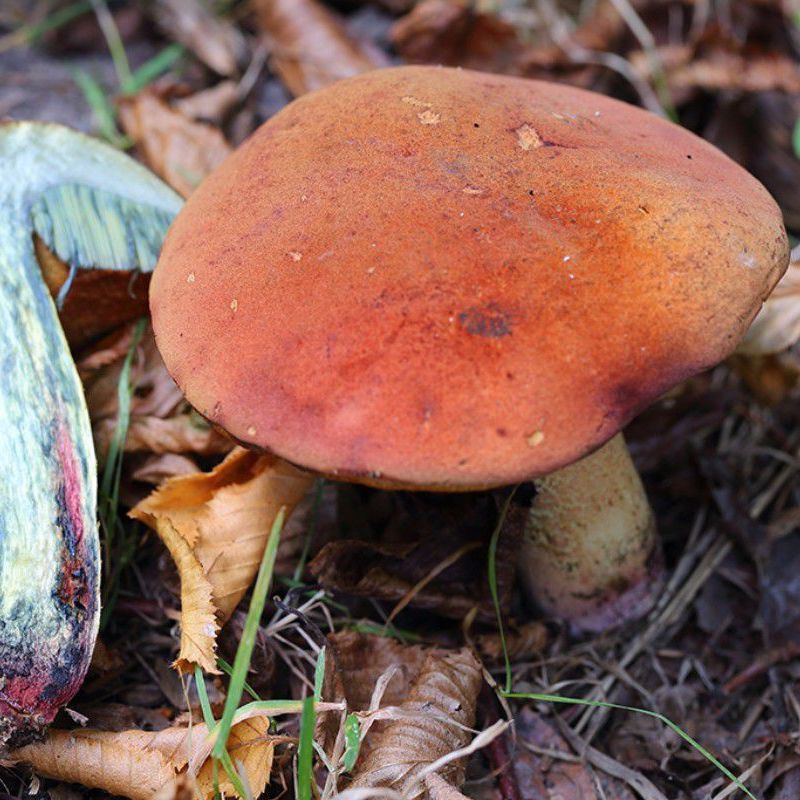} &
    \includegraphics[width=0.105\linewidth, height=0.105\linewidth]{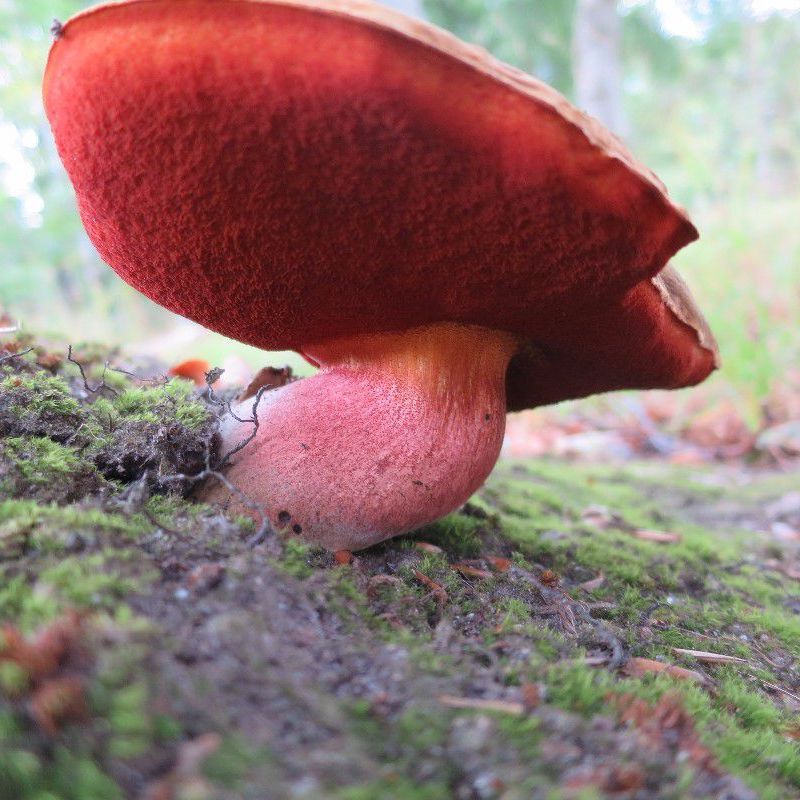} &
    \includegraphics[width=0.105\linewidth, height=0.105\linewidth]{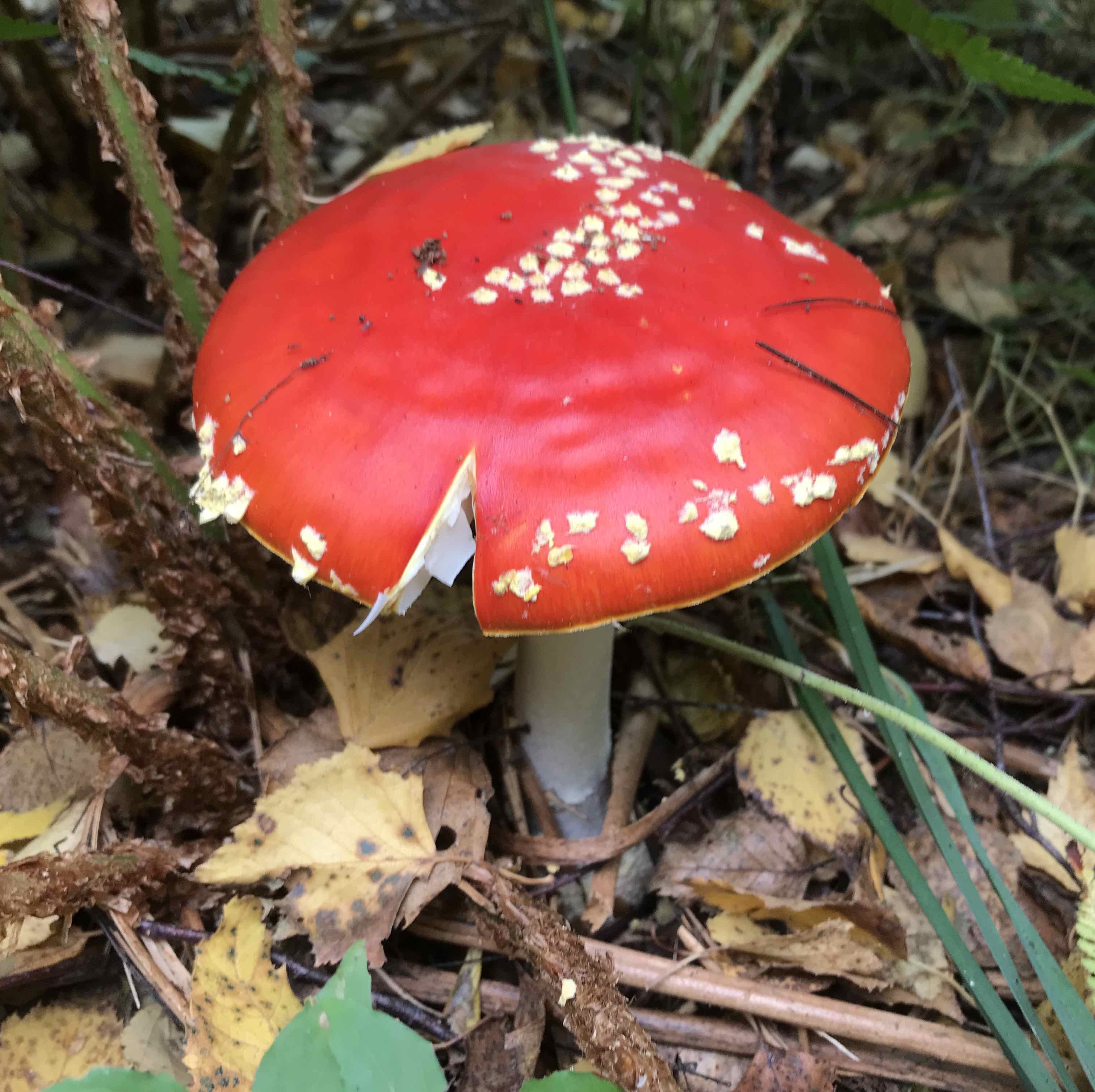} &
    \includegraphics[width=0.105\linewidth, height=0.105\linewidth]{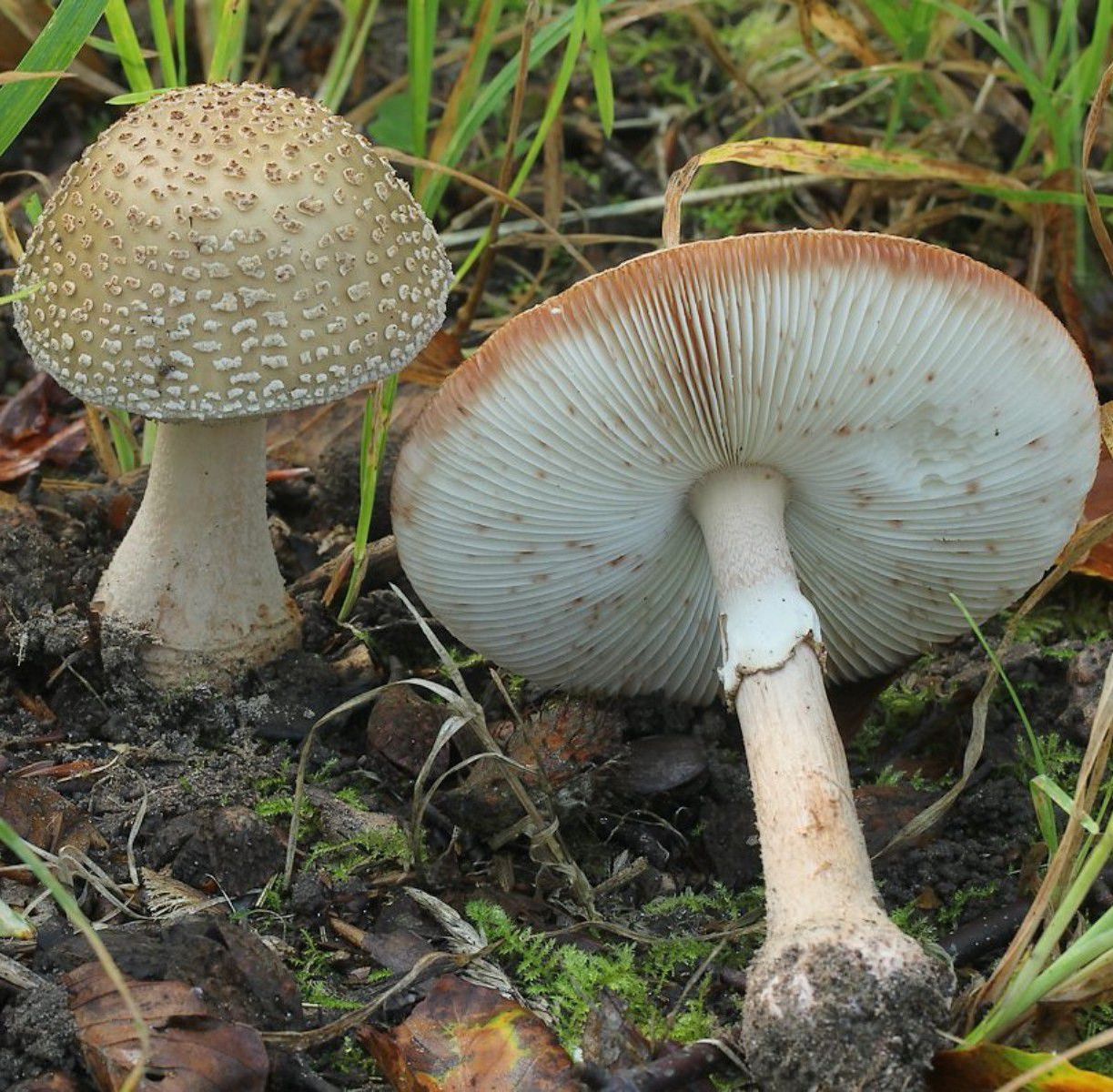} &
    \includegraphics[width=0.1025\linewidth, height=0.1025\linewidth]{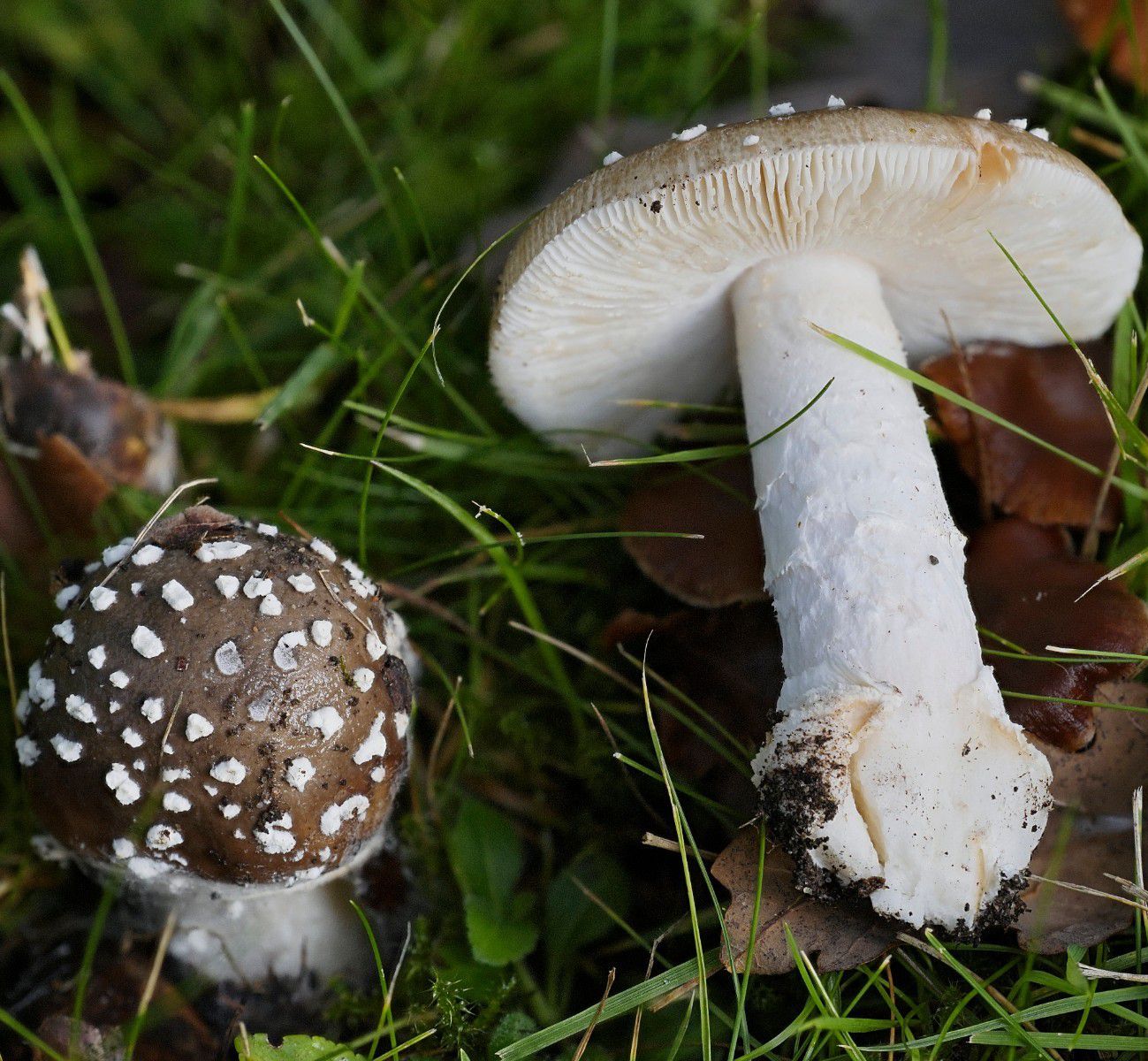} \\
    \hline
  \end{tabular}
  \end{center}
  \caption{Examples of intra- and inter-class similarities and differences  for selected species of three taxonomically distinct fungi families. The similarity holds on the species and the family level. Left: \textit{Russulaceae}, center: \textit{Boletaceae}, right: \textit{Amanitaceae}.}
  \label{fig:fungi_similarity}
\end{figure*}

\section{Related Work}

This section overviews existing fine-grained image datasets,
which, unlike datasets with visually distinct object classes \cite{voc, cifar}, are characterized by small inter-class differences and huge intra-class similarity.
Currently, there exists a number of FGVC dataset with a focus on plants\,\cite{plantclef2017, vegfru_dataset, plantclef2015, leafsnap, dataset-flower}, animals\,\cite{dataset-birdsnap, dataset-dataset-DOGS, nabirds_dataset, dataset-CUBS}, cars\,\cite{dataset-Cars,vmmrdb} or airplanes\,\cite{aircafts}. 
The dataset statistics are compared in Table\,\ref{table:fgvc_datasets}. Most of the datasets are artificially constructed to have a flat class distribution. Many datasets use web scraped data that may contain out-of-domain images and wrong labels.

Fungi species have been covered by image classification datasets.
In the FGVCx 2018 Fungi classification challenge\,\footnote{\url{https://github.com/visipedia/fgvcx_fungi_comp}}, a dataset sampled from the same source as DF20 was used.
The challenge dataset was smaller, scrambled the species names, did not include taxonomic labels and did not contain any metadata.
The latest edition of iNaturalist dataset includes 90,048 images of 384 fungi species from 240 genera.
DF20 is more fine-grained in the \textit{Fungi} kingdom, and it is thus more challenging, with more than 1,500 fungi species. Many of these visually similar but from different genera or families\,(Figure\,\ref{fig:fungi_similarity}).

\textbf{Labels.} As species-level labels are essential for usage in real-world applications, the tedious labelling procedure often rely on domain experts. With just a small number of experts and their limited time, the labelling process is frequently delegated to crowd-sourced annotation platforms such as the Amazon Mechanical Turk\,\cite{imagenet, dataset-dataset-DOGS, dataset-CUBS}. The main drawback of this approach is related to poor domain knowledge of the annotators that results in a high number of noisy labels\,\cite{nabirds_dataset} -- 4.4\% in CUB\,200-2011  and approximately 4.0\% for fine-grained classes in ImageNet. To address this issue, more recent datasets use citizen-science platforms and their users -- citizen scientists\footnote{\,Domain specific nonprofessional enthusiasts - experts.} -- to label data with high-quality annotations\,\cite{nabirds_dataset, inaturalist2017}.

\textbf{ImageNet Overlap.} Strict separation of training and test data is a core machine learning principle and it is standard in the field of image recognition.
Nevertheless, some datasets containt ImageNet images in their test set, and thus
fine-tuning from ImageNet weights contradicts the separation principle. This is commonly overlooked and may lead to biased (inflated) test set accuracies.
For instance, a number of publications with high impact used ImageNet weights and performed the fine-tuning and testing with the CUB\,200-2011\,\cite{dataset-CUBS} dataset that overlaps with the ImageNet\,\cite{ft_imagenet_7, ft_imagenet_2, ft_imagenet_4, ft_imagenet_3, ft_imagenet_5, ft_imagenet_6, ft_imagenet_8, ft_imagenet_9, ft_imagenet_1} in 43 out of 5,794 images (0.75\%).

\textbf{Metadata.} Besides images and class labels, image classification datasets often provide additional metadata, such as higher taxon labels\,\cite{plantclef2015, inaturalist2017}, label hierarchy\,\cite{imagenet, vegfru_dataset, aircafts}, object parts and attribute annotations\,\cite{plantclef2015, nabirds_dataset, dataset-CUBS}, masks\,\cite{dataset-CUBS}, GPS location\,\cite{plantclef2015}, and time of observation\,\cite{plantclef2015}. The existence of such metadata enables the usage of these datasets in machine learning research beyond image classification. For example\,\cite{bargoti2016image, ellen2019improving, Aodha_2019_ICCV} use location context, and \,\cite{goo2016taxonomy} use taxonomy labels.

\section{Atlas of Danish Fungi}

The Atlas of Danish Fungi (Svampeatlas)\,\cite{svampe_databasen, svampeatlas_data, heilmann2019citizen} is supported by more than 3,300 volunteers who have contributed more than 950,000 content-checked observations of fungi,
many with expert-validated class labels, submitted mostly since 2009.

The project has resulted in a vastly improved knowledge of fungi in Denmark\,\cite{heilmann2019citizen}. 
More than 180 species belonging to \textit{Basidiomycota}\,\footnote{\,a group of fungi that produces their sexual spores (basidiospores) on a club-shaped spore-producing structure (basidium).} have been added to the list of known Danish species, and several species that were considered extinct have been re-discovered. Simultaneously, several search and assistance functions have been developed that present features relating to the individual species, making it much easier to include an understanding of endangered species in nature management and decision-making.

Expert-validated Svampeatlas records are published in the Global Biodiversity Information Facility (GBIF), weekly, since 2017. As of end of July 2021, GBIF included 438,872 such images.

\subsection{Annotation Process}

The Atlas of Danish Fungi uses an interactive labelling procedure for all submitted observations. When a user submits a fungal sighting (record) at species level, a "reliability score" (1--100) is calculated based on following factors:
\begin{itemize}[noitemsep,topsep=0pt,leftmargin=0.5cm]
    \item Species rarity, \ie its relative frequency in the Atlas.
    \item The geographical distribution of the species.
    \item Phenology of the species, its seasonality.
    \item User's historical species-level proposal precision.
    \item As above, within the proposal's higher taxon rank.
\end{itemize}
Subsequently, other users may agree with the proposed species identity, increasing the identification score following the same principles, or proposing alternative identification for non-committal suggestions.  Once the submission reaches a score of 80, the label (identification) is internally approved. Simultaneously, a small group of taxonomic experts (validators) monitor most of the observation on their own. These have the power to approve or reject species identifications regardless of the score in the interactive validation. 
Since 2019, the Atlas of Danish Fungi observation identification has been  streamlined thanks to an image recognition system\,\cite{fungiUsecase}.

\begin{figure}[!t]
  \begin{center}
  \footnotesize
  \renewcommand{\arraystretch}{1.0}
  \setlength{\tabcolsep}{1.5pt}
  \begin{tabular}{|cc|cc|}
  \hline
      \textit{Fungi} & \textit{Protozoa} & \textit{Fungi} & \textit{Chromista} \\
    \includegraphics[width=0.24\linewidth, height=0.24\linewidth]{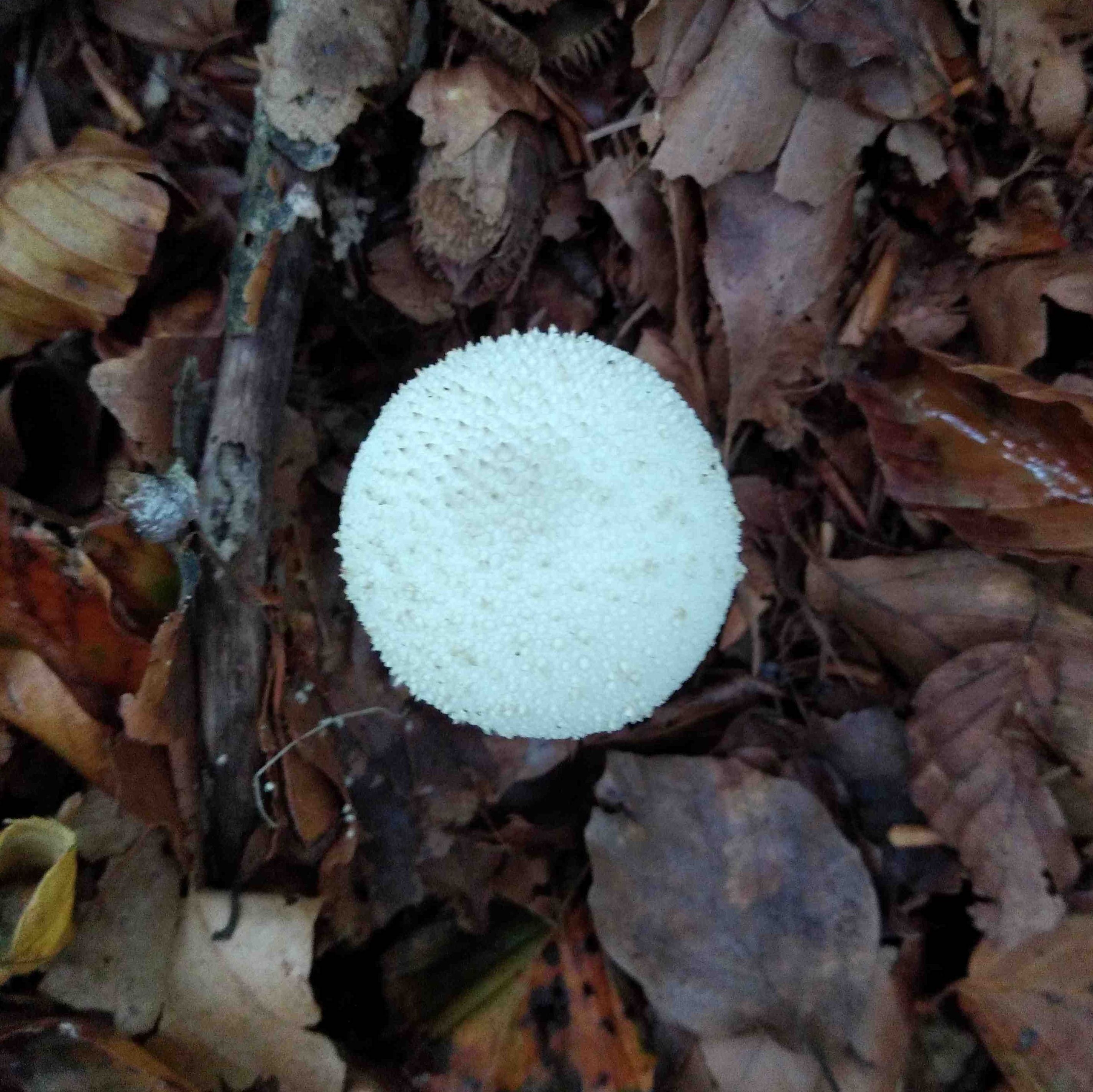} &
    \includegraphics[width=0.24\linewidth, height=0.24\linewidth]{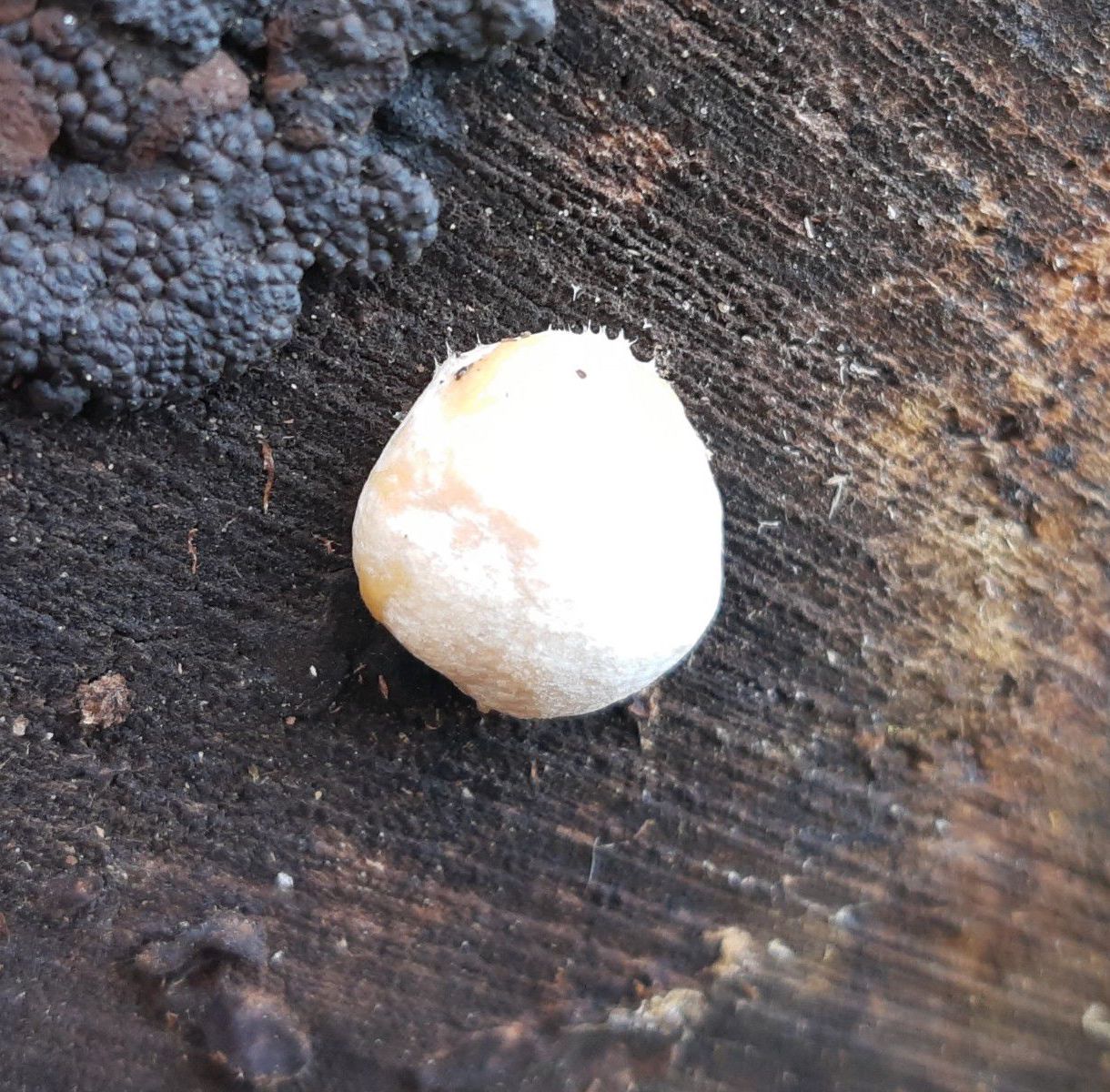} &
    \includegraphics[width=0.24\linewidth, height=0.24\linewidth]{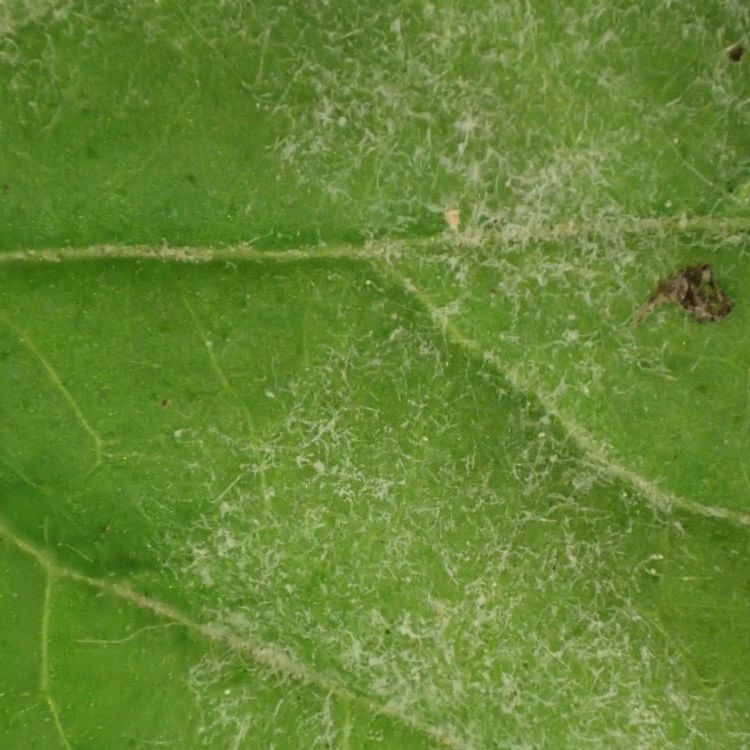} &
    \includegraphics[width=0.24\linewidth, height=0.24\linewidth]{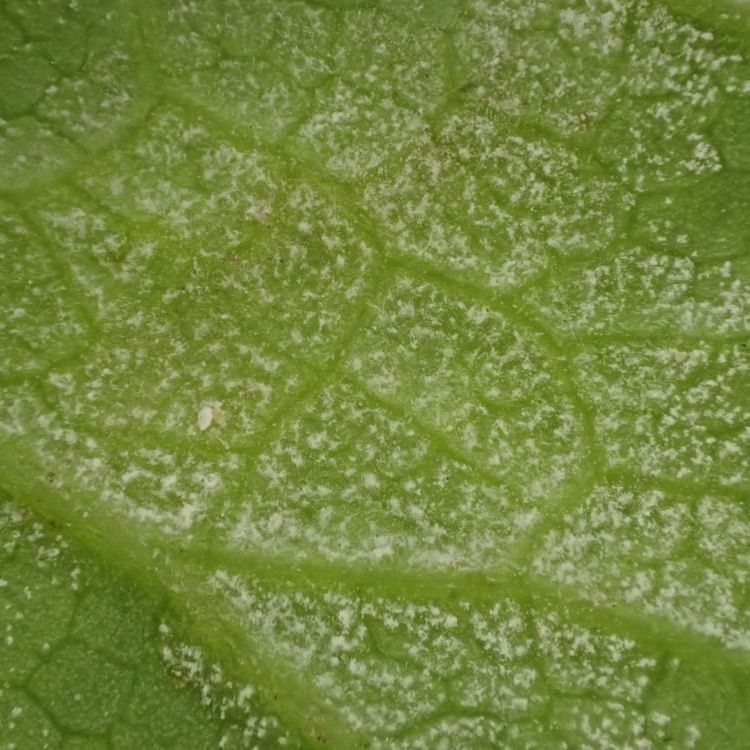} \\
\hline
  \end{tabular}
  \end{center}
  \caption{Visually similar image pairs from the \textit{Fungi} and \textit{Protozoa}, and from the \textit{Fungi} and \textit{Chromista} kingdoms, respectively.}
  \label{fig:fungi_lookalikes}
\end{figure}

\section{Dataset Description}

The \textbf{Danish Fungi 2020} (DF20) dataset contains image observations from the Atlas of Danish Fungi belonging to species with more than 30 images. 
The data are observations collected before the end of 2020\,\footnote{Including 3 preserved specimens collected in 1874, 1882, and 1887, recently photographed.},  originating from 30 countries, and including samples from all seasons. 
It consists of 295,938 images belonging to 1,604 species mainly from the \textit{Fungi} kingdom with  a few visually similar species (See Figure\,\ref{fig:fungi_lookalikes}) from \textit{Protozoa} (1.7\% classes / 1.1\% images) and \textit{Chromista} kingdoms (0.06\% classes / 0.03\% images) kingdoms, covering 566 genera and 190 families.
The most frequent species -- \textit{Trametes versicolor} -- is represented by 1,913 images and the least present with~31.

\begin{table}[t]
\small
\begin{center}
\renewcommand{\arraystretch}{1.1}
\setlength{\tabcolsep}{0.4em}
\begin{tabular}{|l|r|r|r|r|}
\cline{2-5}
\multicolumn{1}{l|}{ } & \multicolumn{1}{c|}{\,\,\,\,Images} & \multicolumn{1}{c|}{\,Species} & \multicolumn{1}{c|}{\,Genera} & \multicolumn{1}{c|}{\,Families} \\
\hline
Svampeatlas (GBIF)& 438,872 & 6,347 & 1,519 & 398 \\
DF20           & 295,938 & 1,604 &   566 & 190 \\
DF20 - Mini    &  36,393 &   182 &     6 &   6 \\
\hline
\end{tabular}
\end{center}
\caption{Numbers of images, species, genera and families in the Atlas of Danish Fungi and their subsets DF20 and DF20 - Mini.}
\label{table:dataset_stats}
\end{table}
Additionally, we hand-picked a subset of 36,393 images belonging to 182 species from 6 genera with a similar visual appearance. This compact dataset, \textbf{DF20\,-\,Mini}, introduces a challenging fine-grained recognition task, while allowing to decrease the necessary training times and hardware requirements.
As species in the same genus are most likely to be confused, we chose six commonly known genera of fungi forming fruit-bodies of the toadstool type with a large number of species: (\textit{Russula}, \textit{Boletus}, \textit{Amanita}, \textit{Clitocybe}, \textit{Agaricus} and \textit{Mycena}) for the construction of the DF20\,-\,Mini. The most frequent species in the DF20\,-\,Mini dataset -- \textit{Mycena galericulata} -- has 1,221 images, the rarest  have 31 samples. For a quantitative summary of the data selection, see Table\,\ref{table:dataset_stats}.

The DF20 and DF20\,-\,Mini datasets were randomly split -- with respect to the class distribution -- into the provided training and (public) test sets, where the training set contains $\lceil 90 \% \rceil$ of images of each species.

\begin{figure*}[t]
\begin{center}
\includegraphics[width=1.0\linewidth]{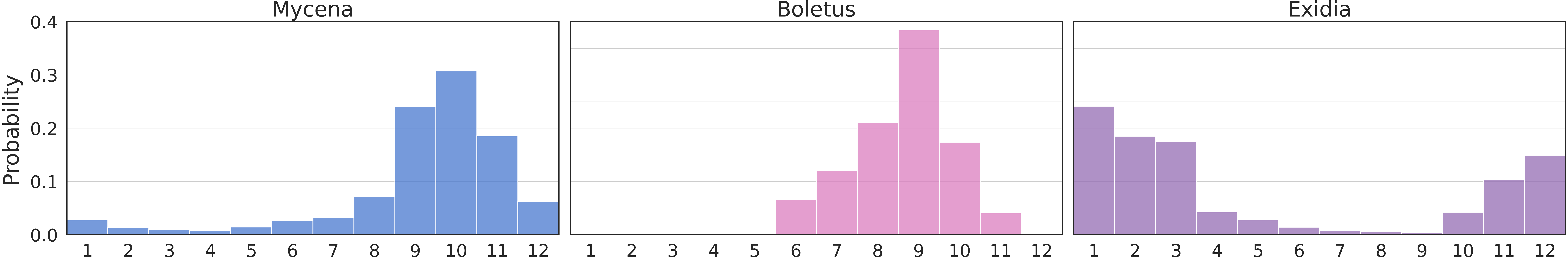}
\caption{Monthly distribution of observations in the DF20 dataset for genera Mycena, Boletus, and Exidia. The differences imply that the class prior distribution varies significantly over time.}
\label{fig:genus_occurence}
\end{center}
\end{figure*}

\subsection{Metadata}
Unlike most computer vision datasets, DF20 and DF20\,-\,Mini include rich metadata acquired by citizen-scientists in the field while recording the observations. 
We see a promising research direction in combining visual data with metadata like timestamp, location at multiple scales, substrate, habitat, full taxonomy labels and camera device settings. For a detailed list see Table\,\ref{table:metadata}. 

\textbf{Substrate.} Substrates on which fungi live and fruit are an important source of information that helps differentiate similarly looking species. Each species or genus has its preferable substrate, and it is rare to find it on other substrates. For example, \textit{Trametes} occurs only on wood and \textit{Russula} on soil. As such metadata is crucial for the final categorization capability, we provide one of 32 substrate types for more than 99\% of images. We differentiate wood of living trees, dead wood, soil, bark, stone, fruits, mosses and others. 

\textbf{Location.} Fungi are highly location dependent with different species distributions across continents, states, regions or even districts. To support studies on better understanding where Fungi species lives, we include multi-level location information. Starting from latitude and longitude values and their uncertainty, we further extracted information about the country, region, district, and locality with 9003 unique values. 

\textbf{Time-Stamp.} Time of observation is essential for fungi classification in the wild as fruitbodies' presence depends on  seasonality or even (but rarely) the time in a day. Considering the existence of such dependency, integrating information about time into the classification should also improve fungal recognition. In Figure\,\ref{fig:genus_occurence} we show the probability of three genera being observed in different months of the calendar year. Brief inspection shows that there is almost zero probability to spot a \textit{Boletus} in January but still a small chance to find a \textit{Mycena}. In contrast to \textit{Boletus}, \textit{Exidia} occurs mostly during the cold months.

\textbf{EXIF data.} Since the camera device and its settings affect the resulting image, the image classification models may be biased towards certain (e.g. more common) device attributes. To allow a deeper study of such phenomena, we include the EXIF data for approximately 84\% of images, where the EXIF information was available in the Atlas of Danish Fungi. The included attributes, the number of unique values in the dataset and the proportion of images with the attributes present are summarized in Table \ref{table:exif_values}. \\

\begin{table}[b!]
\small
\vspace{-0.2cm}
\begin{center}
\renewcommand{\arraystretch}{1.1}
\setlength{\tabcolsep}{0.4em}
\begin{tabular}{|l|c|r|}
\hline
\textbf{Attribute} & \textbf{~~Coverage}\,[\%]~~ & \textbf{~~\# Values~~} \\
\hline
White Balance                 & 79.99 &     2 \\
Color Space                   & 84.38 &     3 \\
Metering Mode                 & 78.23 &    10 \\
SceneCaptureType              & 81.04 &    13 \\
Compressed Bits Per Pixel\,\, & 37.75 &    88 \\
Aperture Value                & 46.63 &   297 \\
Device                        & 80.31 &   688 \\
Focal Length                  & 80.13 & 1,580 \\
Exposure Time                 & 80.12 & 4,594 \\
Shutter Speed Value           & 46.78 & 7,079 \\
\hline
\end{tabular}
\end{center}
\caption{Device settings extracted from the original image EXIFs in the Atlas of Danish Fungi, with the proportion of images where the attributes are present (Coverage), and the number of unique values of the attribute in the dataset.}
\label{table:exif_values}
\end{table}

\textbf{Habitat.} While substrate denotes the spots, the habitat indicates the more overall environment where fungi grow and hence is vital for fungal recognition. It is well known that some species occur in deciduous forests rather than in conifer forests or plantations, while others grow in farmland. For a deeper understanding of such relation, we include the information about the habitat for approximately 99.5\% of observations.

\begin{table}[b!]
\begin{center}
\vspace{-0.2cm}
\setlength{\tabcolsep}{0.28em} 
\begin{tabular}{| r  p{6.0cm}@{\hskip 5pt}  |}
    \hline
    \small{\textbf{Attribute}} & \small{\textbf{Description}} \\
    \hline
        \footnotesize{\textbf{EventDate}} & \footnotesize{Date of observation.} \\
        \footnotesize{\textbf{EXIF}} & \footnotesize{Camera device attributes extracted from the image, e.g., metering mode, color space, device type, exposure time, and shutter speed.} \\
        \hline
        \footnotesize{\textbf{Habitat}} & \footnotesize{The environment where the specimen was observed. Selected from 32 values such as Mixed woodland, Deciduous woodland etc.}\\
        \footnotesize{\textbf{Substrate}} & \footnotesize{The natural substance on which the specimen lives. A\,total of 32 values such as Bark, Soil, Stone, etc.} \\
        \hline
        \footnotesize\textbf{{Scientific}} & \footnotesize{Lowest taxonomic rank including specific Epithets. 1,604 unique values present.} \\
        \footnotesize{\textbf{Species}} & \footnotesize1th taxon rank. 1,578 unique values present. \\ 
        \footnotesize{\textbf{Genus}} & \footnotesize{2nd taxon rank. 566 unique values present.} \\
        \footnotesize{\textbf{Family}} & \footnotesize{3rd taxon rank. 190 unique values present.} \\
        \footnotesize{\textbf{Order}} & \footnotesize{4th taxon rank. 66 unique values present.} \\
        \footnotesize{\textbf{Class}} & \footnotesize{5th taxon rank. 23 unique values present.} \\
        \footnotesize{\textbf{Phylum}} & \footnotesize{6th taxon rank. 5 unique values present.} \\
        \footnotesize{\textbf{Kingdom}} & \footnotesize{7th taxon rank. 3 unique values present.} \\  
      \hline
        \footnotesize{\textbf{CountryCode}} & \footnotesize{ISO 3166-1 alpha-2 code (DK, AT, etc.) of the observation. The dataset covers 30 countries.} \\ 
        \footnotesize{\textbf{Locality}} & \footnotesize{More precise location information. Mostly smaller than a district, e.g. part of a city or a specific forest. 9003 values present.}  \\
        \footnotesize{\textbf{Level1Gid}} & \footnotesize{ID of a Country region related to the specimen observation, 115 regions are listed.} \\
        \footnotesize{\textbf{Level2Gid}} & \footnotesize{ID of a district region related to the specimen observation, 317 districts are listed.} \\
        \footnotesize{\textbf{Latitude}} & \footnotesize{A decimal GPS coordinate.} \\
        \footnotesize{\textbf{Longitude}} & \footnotesize{A decimal GPS coordinate.} \\
        \footnotesize{\textbf{GPSUncert}} & \footnotesize{GPS coordinates uncertainty in meters.} \\
      \hline
\end{tabular}
\end{center}
\caption{Description of the provided metadata (observation attributes). For almost all images, a detailed information about taxonomy, location, time, habitat and substrate type is included.}
\label{table:metadata}
\end{table}

\pagebreak

\section{Experiments}
\label{experiments}

To establish a baseline performance on the DF20 and DF20\,-\,Mini datasets, we performed multiple experiments. First, we train a wide variety of well known CNN architectures such as ResNets\,\cite{resnets}, EfficientNets\,\cite{tan2019efficientnet}, MobileNet\,\cite{sandler2018mobilenetv2}, Inception networks\,\cite{inceptions} and \mbox{SE-ResNeXt-101-32x4d} that extends the ResNet-101 by cardinality\,\cite{next} and Squeeze and Excite blocks\,\cite{SE_Net}. Second, the EfficientNet-B0, EfficientNet-B3, and \mbox{SE-ResNeXt-101-32x4d} are compared with Vision Transformer architectures  ViT-Large/16 and ViT-Base/16\,\cite{vit}. 
Finally, the impact of different metadata and their combinations on both the CNN and the ViT final prediction performance is evaluated.

\subsection{Setup}
\label{setup}

In this section, we describe the full training and evaluation procedure, including the training strategy and image augmentations.

\textbf{Training Strategy.} All architectures were initialized from publicly available ImageNet-1k pre-trained checkpoints and further fine-tuned with the same strategy for 100 epochs with the PyTorch\,framework\,\cite{PyTorch} within the 21.05 NGC deep learning framework Docker container. All neural networks were optimized by Stochastic Gradient Descent 
with momentum 
set to 0.9. The start Learning Rate (LR) was set to 0.01 and was further decreased with a specific adaptive learning rate schedule strategy -- if the validation loss is not reduced from one epoch to another, reduce LR by 10\%. To have the same effective mini-batch size of 64 for all architectures, we accumulated gradients from smaller mini-batches accordingly, where needed.

\textbf{Augmentations.} For training, we utilized several augmentation techniques from the Albumentations\,library\,\cite{albumentation}. More specifically, we used: random horizontal flip with 50\% probability, random vertical flip with 50\% probability, random resized crop with a scale of 0.8 - 1.0, random brightness\,/\,contrast adjustments with 20\% probability, and mean and std. dev. normalization. 
All images were resized to the required network input size: For the CNN performance experiment, inputs of size 299$\times$299 were used. In the case of the CNN vs ViT experiment, we used two different resolutions, 224$\times$224 and 384$\times$384, to match the input resolutions of the pre-trained models.

\textbf{Test-time.} While testing, we avoided any additional techniques such as ensembles, centre-cropping, prior weighting, etc. Only the resize and normalization operations were used to pre-process the data. The impact of test-time augmentation methods on the final performance can be studied in the future.

\subsection{Metadata Use}
\label{metadata_usage}
We propose a simple method for the use of metadata to improve the categorization performance -- similar to spatio-temporal prior used in\,\cite{dataset-birdsnap}.
For a given type of metadata ($D$) and image ($I$), we adopt the following assumption for the likelihood of an image observation: 
\begin{equation}
\small
P(I|S) = P(I|S,D),
\label{eq:assumption}
\end{equation}
i.e. that the visual appearance of a species does not depend on the metadata.
This does not mean that the posterior probability of a species given an image is independent of metadata $D$.

A few lines of algebraic manipulation prove that under assumption Eq.\,\eqref{eq:assumption}, the class posterior given the image $I$ and metadata $D$ is easily obtained:
\begin{equation}
\small
\begin{aligned}
  P(S|I,D)  & =       P(S|I) \frac{P(S|D)}{P(S)}\frac{P(I)}{P(I|D)} \\
            & \propto P(S|I) \frac{p(S|D)}{p(S)},
\label{eq:posterior-given-meta}
\end{aligned}
\end{equation}
where $P(S)$ is the class prior in the training set.
The discrete conditional probability $P(S|D)$ is estimated as the relative frequency of species $S$ with metadata $D$ in the training\,set.

While we know this assumption is not always true in practice, since metadata like substrate or time in fact do impact the image background as well as the appearance of the specimen, 
this is the only possible approach not requiring modelling the dependence of visual appearance and the metadata. 
The model trained without metadata has no information about visual appearance changes of a species as a function of $D$. Moreover, this assumption is applicable for situations where the classifier has to be treated as a black box without the possibility to retrain the model.
Even this simplistic model based on an unrealistic assumption reduces error rates, see
Table~\ref{table:results_metadata}.

With multiple metadata at once, e.g. month and habitat, we combine the posteriors assuming statistical independence:
\begin{equation}
\small
    P (S | D_1,D_2) \propto \frac{P (S | D_1 ) P(S|D_2)}{P(S)}.
\end{equation}
This is a simple, baseline assumption, which again may not always be valid for related meta-data. Direct estimation of $P(S | D_1,D_2)$, e.g. as relative frequencies, is another possibility. The D20 benchmark has thus the potential to be a fertile ground for evaluation of intra-metadata, as well as visual-metadata, dependencies.

The approach of Eq.\,(\ref{eq:posterior-given-meta}) needs a probabilistic classifier to serve as an estimator of $P(S|I)$. In our experiments, we use the outputs of the softmax layer.
Note that for CNNs, the estimates of $\max P(S|I)$ are typically overconfident, and the quality of the estimator can be improved by a process is called {\it calibration} in the literature\,\cite{guo2017calibration,vaicenavicius2019evaluating}. 
The proposed benchmark may be used, in the context of exploiting metadata, to evaluate and compare classifier calibration techniques.

\subsection{Metrics}
\label{metrics}

\begin{table}[b]
\footnotesize
\begin{center}
\setlength{\tabcolsep}{0.3em} 
\renewcommand{\arraystretch}{1.17}
\begin{tabular}{|l| c | c | c || c | c | c |}
\cline{2-7}
    \multicolumn{1}{c|}{ } & \textbf{Top1} & \textbf{Top3} & \,\,\textbf{$\text{F}_{1}^{m}$}\,\, &  \textbf{Top1} & \textbf{Top3} & \,\,\textbf{$\text{F}_{1}^{m}$}\,\,  \\
    \hline
    MobileNet-V2          & \,65.58\, & \,83.65\, & \,0.550\, & \,69.77\, & \,85.01\, & \,0.606\, \\
    \hline 
	ResNet-18             & 62.91 & 81.65 & 0.514 & 67.13 & 82.65 & 0.580 \\
	ResNet-34             & 65.63 & 83.52 & 0.559 & 69.81 & 84.76 & 0.600 \\
	ResNet-50             & 68.49 & 85.22 & 0.587 & 73.49 & 87.13 & 0.649 \\
	\hline
	EfficientNet-B0       & 67.94 & 85.71 & 0.567 & 73.65 & 87.55 & 0.653 \\
	EfficientNet-B1       & 68.35 & 84.67 & 0.572 & 74.08 & 87.68 & 0.654 \\
	EfficientNet-B3       & 69.59 & 85.55 & 0.590 & 75.69 & 88.72 & 0.673 \\
	EfficientNet-B5       & 68.76 & 85.00 & 0.579 & 76.10 & 88.85 & 0.678 \\
	\hline 
	Inception-V3          & 65.91 & 82.97 & 0.535 & 72.10 & 86.58 & 0.630 \\
	Inception-ResNet-V2\, & 64.67 & 81.42 & 0.542 & 74.01 & 87.49 & 0.651 \\
	Inception-V4          & 67.45 & 82.78 & 0.560 & 73.00 & 86.87 & 0.637 \\ 
	\hline
	SE-ResNeXt-101 & \textbf{72.23} & \textbf{87.28} & \textbf{0.620} & \textbf{77.13} & \textbf{89.48} & \textbf{0.693} \\ 
	\hline	
 \multicolumn{1}{c}{ } & \multicolumn{3}{|c||}{DF20 - Mini} & \multicolumn{3}{c|}{DF20} \\
 \cline{2-7}
\end{tabular}
\end{center}
\caption{Classification performance of selected CNN architectures on DF20 and DF20\,-\,Mini. All networks share the settings described in Section \ref{setup} and were trained on 299$\times$299 images. The top results  -- $\text{F}_{1}^{m}$, see Eq. (\ref{eq:F1_macro}), equal to 0.620 / 0.693 and Top1 to 72.23\% / 77.13\% -- are far from saturated. The datasets are challenging for the state-of-the-art CNN classifiers.}
\label{table:results}
\end{table}

Besides commonly used metrics, Top1 and Top3 accuracy, we measured the macro-averaged $\text{F}_1$ score, $\text{F}_{1}^{m}$, which is not biased by class frequencies and is more suitable for the long-tailed class distributions observed in the nature. Interestingly, even though the performance across the whole taxonomy in nature-related FGVC datasets is highly demanded, most existing datasets are only using accuracy as the score measure. Considering that the datasets are highly imbalanced with long-tail distribution, learning procedure may ignore the least present species. Additionaly, usage of $\text{F}_{1}^{m}$ allows to easily assign a cost value to both types of error\,({\it fp} and {\it fn}) for each label and to measure more task-relevant performance. For example, in fungi recognition, mistaking a poisonous mushroom for the edible one is a more significant problem than the opposite. 

The $\text{F}_{1}^{m}$ is defined as the mean of class-wise $\text{F}_1$ scores:

\begin{equation}
\small
\text{F}_{1}^{m} = \frac{1}{N} \sum_{S=1}^{N} F_{1_{S}}\,,
\label{eq:F1_macro}
\end{equation}

where\,$N$ represents the number of classes and $S$ is the species index. Than the $\text{F}_1$ score for each class is calculated as a harmonic mean of the class precision $P_S$ and recall $R_S$:

\begin{equation}
\small
F_{1_{S}} = 2 \times \frac{{P_S} \times {R_S}}{P_S + R_S}\,,
\label{eq:F1_i}
\end{equation}

\begin{equation}
\small
P_S=\frac{\text{\it tp}_S}{\text{\it tp}_S+\text{\it fp}_S}\,,\,R_S=\frac{\text{\it tp}_S}{\text{\it tp}_S+\text{\it fn}_S}\,
\label{eq:precisionrecall}
\end{equation}

In multi-class classification, the True Positive\,({\it tp}) represents the number of correct Top1 predictions, False Positive ({\it fp}) how many times was a specific class predicted instead of the\,({\it tp}), and False Negative\,({\it fn}) how many images of class $S$ have been misclassified.

\subsection{Results}
\label{results}

In this section, we compare the performance of the well known CNN based models and ViT models in terms of Top1 and Top3 accuracy, and the newly included $\text{F}_{1}^{m}$ metric. Additionally, we discuss the impact of the metadata on the classification performance and differences in performance between CNNs and ViTs.

\begin{table}[b]
\footnotesize
\begin{center}
\setlength{\tabcolsep}{0.3em} 
\renewcommand{\arraystretch}{1.17}
\begin{tabular}{|l| c | c | c || c | c | c | c |}
    \cline{2-7}
    \multicolumn{1}{c|}{ } & \textbf{Top1} & \textbf{Top3} & \,\,\textbf{$\text{F}_{1}^{m}$}\,\, & \textbf{Top1} & \textbf{Top3} & \,\,\textbf{$\text{F}_{1}^{m}$}\,\, & \multicolumn{1}{c}{ }\\  
    \hline
    EfficientNet-B0     & \,\,65.66\,\, & \,\,83.65\,\, & \,0.531\, & \,\,70.33\,\, & \,\,85.19\,\, & \,0.613\, & \multirow{5}{*}{\rotatebox[origin=c]{90}{$224\times224$}} \\
	EfficientNet-B3     & 67.39 & 83.74 & 0.550 & 72.51 & 86.77 & 0.634  & \\
	SE-ResNeXt-101      & 68.87 & 85.14 & 0.585 & 74.26 & 87.78 & 0.660 & \\
	ViT-Base/16         & 70.11 & \textbf{86.81} & 0.600 & 73.51 & 87.55 & 0.655 & \\
	ViT-Large/16        & \textbf{71.04} & 86.15 & \textbf{0.603} & \textbf{75.29} & \textbf{88.34} & \textbf{0.675} & \\
	\hline
	\noalign{\vskip 0.5mm}
	\hline
    EfficientNet-B0     & 69.62 & 85.96 & 0.582 & 75.35 & 88.67 & 0.670  & \multirow{5}{*}{\rotatebox[origin=c]{90}{$384\times384$}} \\
    EfficientNet-B3     & 71.59 & 87.39 & 0.613 & 77.59 & 90.07 & 0.699 & \\
    SE-ResNeXt-101      & 74.23 & 88.27 & 0.651 & 78.72 & 90.54 & 0.708 & \\
    ViT-Base/16         & 74.23 & 89.12 & 0.639 & 79.48 & 90.95 & 0.727 & \\
	ViT-Large/16        & \textbf{75.85} & \textbf{89.95} & \textbf{0.669} & \textbf{80.45} & \textbf{91.68} & \textbf{0.743} & \\
	\hline
	 \multicolumn{1}{c}{ } & \multicolumn{3}{|c||}{DF20 - Mini} & \multicolumn{3}{c|}{DF20}  & \multicolumn{1}{c}{ }\\  
	  \cline{2-7}
\end{tabular}
\end{center}
\caption{Classification results of selected CNN and ViT architectures on DF20 and DF20\,-\,Mini datasets. ViT  achieves  results  superior to CNN baselines with 80.45\% accuracy and 0.743 $\text{F}_{1}^{m}$, reducing the CNN error by 9\% and 12\% respectively.}
\label{table:results_ViT}
\end{table}

\textbf{Convolutional Neural Networks.} 
Comparing well known CNN architectures on DF20 and DF20\,-\,Mini, we can see a similar behaviour as on other datasets\,\cite{imagenet, inaturalist2017, dataset-CUBS}. 
The best performing model on both datasets was SE-ResNeXt-101 with 0.620 $\text{F}_{1}^{m}$ score on DF20\,-\,Mini and 0.693 $\text{F}_{1}^{m}$ score on DF20. A more detailed comparison of the achieved scores (Top1, Top3, and $\text{F}_{1}^{m}$) for each model are summarized in Table\,\ref{table:results}. 

\textbf{Vision Transformers.}
The recently introduced ViT\,\cite{vit} showed excellent performance in object classification compared to state-of-the-art convolutional networks. Apart from the CNNs, the ViT is not using convolutions but interprets an image as a sequence of patches and process it by a standard Transformer encoder as used in  natural language processing\,\cite{vaswani2017attention}.
To evaluate its performance for transfer-learning in the FGVC domain, we compare two ViT architectures - ViT-Base/16 and ViT-Large/16 - against the well performing CNN models - EfficientNet-B0, EfficientNet-B3 and SE-ResNeXt-101. As ImageNet pre-trained ViT models were available just for input sizes of 224$\times$224 and 384$\times$384, we trained all networks on these resolutions while following the training setup fully described in subsection \ref{setup}. Differently from the performance validation of Dosovitskiy et al.\,\cite{vit} on ImageNet, in our evaluation on DF20, ViTs ourperform state-of-the-art CNNs by a large margin. The best performing ViT model achieved an impressive 0.743 $\text{F}_{1}^{m}$ score while outperforming the SE-ResNeXt-101 by a significant margin of 0.035 in $\text{F}_{1}^{m}$, and 1.73\% of Top1 accuracy on the images with 384$\times$384 input size. In the case of the  224$\times$224, we see a smaller margin of 1.62\% in Top1 accuracy and 0.018 in the  $\text{F}_{1}^{m}$ score. Wider performance comparison is shown in Table\,\ref{table:results_ViT}.


\begin{table}[t!]
\footnotesize
\begin{center}
\setlength{\tabcolsep}{0.4em} 
\renewcommand{\arraystretch}{1.17}
\begin{tabular}{| c | c | c | c | c | c || c | c | c |}
\hline
    \textbf{H} & \textbf{M} & \textbf{S} & \textbf{Top1} & \textbf{Top3} & \textbf{$\text{F}_{1}^{m}$}  & \textbf{Top1} & \textbf{Top3} & \textbf{$\text{F}_{1}^{m}$} \\
    \hline
     $\times$   & $\times$   & $\times$   & \,\,80.45\,\, & \,\,91.68\,\, & ~\,\,\,\,0.743\,\,  & \,\,73.51\,\, & \,\,87.55\,\, & ~\,\,\,\,0.655\,\,    \\
     \hline
     \faCheck{} & $\times$   & $\times$   & +1.50 & +1.00 & +0.027 & +1.94 & +1.50 & +0.040   \\
     $\times$   & \faCheck{} & $\times$   & +0.95 & +0.62 & +0.014 & +1.23 & +0.95 & +0.020   \\
     $\times$   & $\times$   & \faCheck{} & +1.13 & +0.69 & +0.020 & +1.39 & +1.17 & +0.025   \\
     $\times$   & \faCheck{} & \faCheck{} & +1.93 & +1.27 & +0.032 & +2.47 & +1.98 & +0.042   \\
     \faCheck{} & $\times$   & \faCheck{} & +2.48 & +1.66 & +0.044 & +3.23 & +2.47 & +0.062   \\
     \faCheck{} & \faCheck{} & $\times$   & +2.31 & +1.48 & +0.040 & +3.11 & +2.30 & +0.057   \\
     \hline
     \faCheck{} & \faCheck{} & \faCheck{} & +\textbf{2.95} & +\textbf{1.92} & +\textbf{0.053} & +\textbf{3.81} & +\textbf{2.84} & +\textbf{0.070}  \\
     \hline
     \multicolumn{1}{}{} & \multicolumn{1}{}{} & \multicolumn{1}{c|}{} & \multicolumn{3}{c||}{ViT-Large/16 -- $384\times384$} & \multicolumn{3}{c|}{ViT-Base/16 -- $224\times224$}   \\
     \cline{4-9}
\end{tabular}
\end{center}
\caption{Performance gains based on 3 observation metadata and their combination. DF20. \textbf{H}\,-\,Habitat, \textbf{S}\,-\,Substrate, \textbf{M}\,-\,Month.}
\label{table:results_metadata}
\end{table}

\textbf{Importance of the metadata.} Inspired by the common practice in mycology, we set up an experiment to show the importance of metadata for \textit{Fungus} species identification. Using the approach described in Section\,\ref{metadata_usage}, we improved performance in all measured metrics by a significant margin. We measured the performance improvement with all metadata types and their combinations. Overall, habitat was most efficient in improving the performance. With the combination of habitat, substrate and month, we improved the ViT-Large/16 model's performance on DF20 by 2.95\%, 1.92\% and 0.053 in Top1, Top3 and $\text{F}_{1}^{m}$, respectively, and the performance of the ViT-Base/16 model by 3.81\%, 2.84\% and 0.070 in Top1, Top3 and $\text{F}_{1}^{m}$
Detailed evaluation of the performance gain using different observation metadata and their combinations is shown in Table\,\ref{table:results_metadata}.

\textbf{DF20 vs DF20\,-\,Mini}. The performance evaluation with selected CNN and ViT architectures showed that even with a smaller number of classes and one-tenth of the data, DF20\,-\,Mini as a compact subset of DF20 offers an even more challenging problem for state-of-the-art architectures while being less time and hardware demanding.

\section{Conclusion}

This paper introduced a novel fine-grained dataset and classification benchmark, the Danish Fungi 2020, and its subset, Danish Fungi 2020 - Mini. 
The dataset was constructed from data submitted to the Atlas of Danish Fungi and labeled by mycologists. It includes 295,938 photographs of 1,604 species -- mainly from the Fungi kingdom -- together with full taxonomic labels, rich metadata, compact size and severe difficulty, and the same training and test set species distribution.

The quantitative and qualitative analysis of CNNs and ViTs shows superior performance of the ViT in fine-grained classification. We present the baselines for processing the habitat, substrate and time (month) metadata. We show that -– even with the simple method from Section \ref{metadata_usage} –- utilizing the metadata increases the classification performance significantly. We provide the code and trained model check-points to all our baselines. 
A publicly available web benchmark allows -- through CSV submision file -- for an on-line comparison of state-of-the-art results for both image-only and image\,+\,metadata submissions. With the precise annotation and rich metadata, we would like to encourage further research in other areas of computer vision and machine learning, beyond fine-grained visual categorization. The benchmark may help research in classifier calibration, loss functions, validation metrics, taxonomy\,/\,hierarchical learning, device dependency or time series based species prediction. For example, the standard loss function focusing on recognition accuracy ignores the practically important cost of predicting a species with high toxicity. 

\begin{figure}[t!]
\begin{center}
\includegraphics[width=1.0\linewidth]{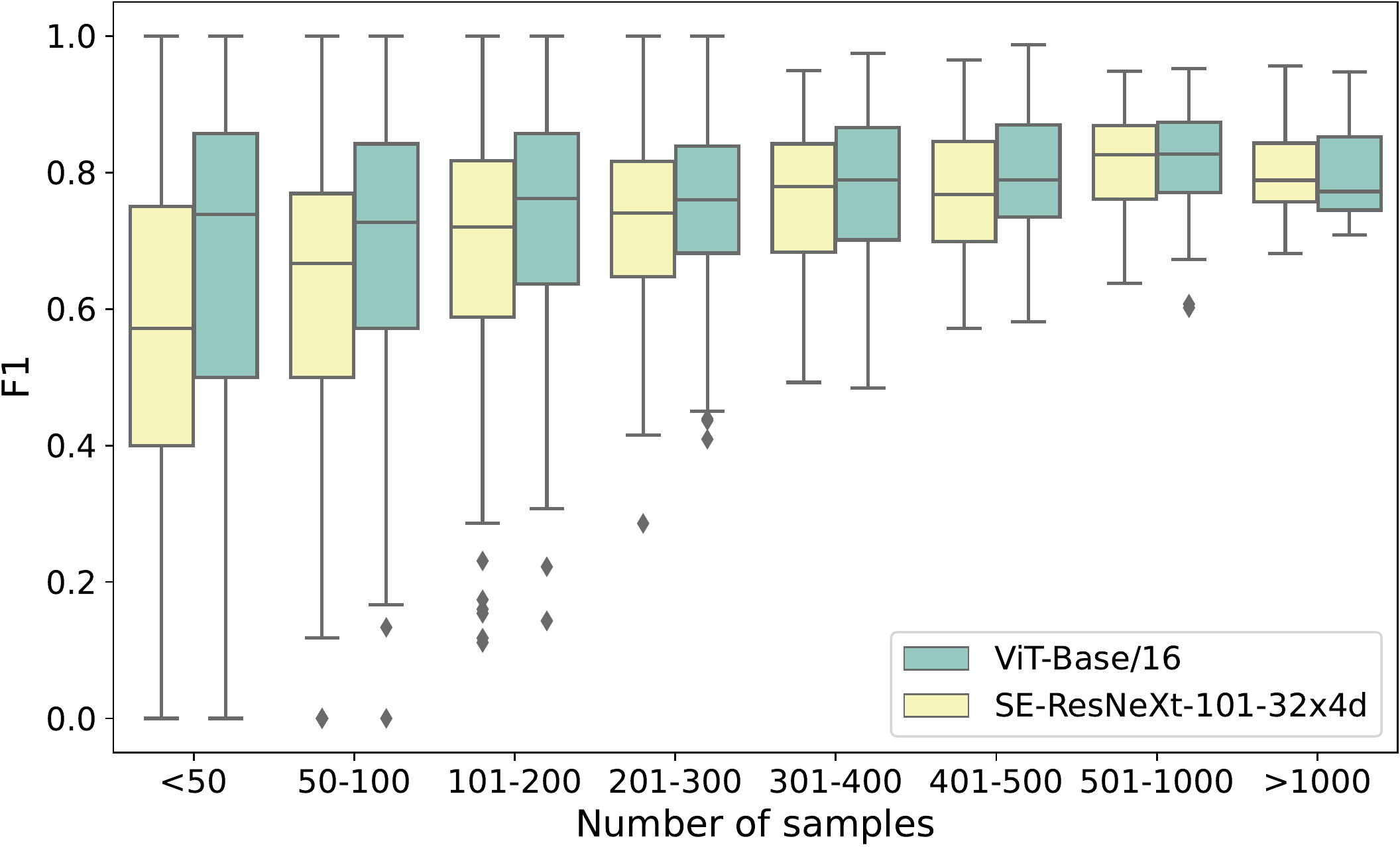}
\caption{Box plot of the dependence of classification performance ($\text{F}_{1}$) on the number of training samples of a class. Tested on DF20 with input resolution of $224\times224$.}
\label{fig:perf_comp}
\end{center}
\end{figure}




{\small
\bibliographystyle{ieee_fullname}
\bibliography{paper}
}

\end{document}